%% file: bare_jrnl_new_sample4.tex
\documentclass[lettersize,journal]{IEEEtran}
\usepackage{amsmath,amsfonts}
\usepackage{algorithmic}
\usepackage{algorithm}
\usepackage{array}
\usepackage[caption=false,font=normalsize,labelfont=sf,textfont=sf]{subfig}
\usepackage{textcomp}
\usepackage{stfloats}
\usepackage{url}
\usepackage{verbatim}
\usepackage{graphicx}
\usepackage{cite}
\usepackage[table]{xcolor}
\usepackage{multirow}
\usepackage{adjustbox}
\usepackage{booktabs}
\usepackage{arydshln}
\usepackage{pifont}
\usepackage{xcolor}
\usepackage{pgfplots}
\pgfplotsset{compat=1.18}  
\usepackage{tikz}
\usetikzlibrary{plotmarks}   
\usetikzlibrary{positioning} 

\usepackage[hidelinks, colorlinks=true, linkcolor=blue, citecolor=blue, urlcolor=blue]{hyperref}
\definecolor{mygreen}{HTML}{97C04B}
\definecolor{myorange}{HTML}{E0713F}
\definecolor{numbergreen}{HTML}{32CD32}
\newcommand{\revision}[1]{#1}

\newcommand{\graycell}[1]{\textcolor{lightgray}{#1}}

\begin{document}

\title{SPFSplatV2: Efficient Self-Supervised Pose-Free 3D Gaussian Splatting from Sparse Views}

\author{Ranran Huang, Krystian Mikolajczyk
\thanks{Ranran Huang, and Krystian Mikolajczyk are with the Department
of Electrical and Electronic Engineering, Imperial College London, London SW7 2AZ, United Kingdom. E-mail: \{r.huang24; 
k.mikolajczyk\}@imperial.ac.uk.}
}

\markboth{Journal of \LaTeX\ Class Files,~Vol.~14, No.~8, August~2025}%
{Shell \MakeLowercase{\textit{et al.}}: A Sample Article Using IEEEtran.cls for IEEE Journals}

\IEEEpubid{0000--0000/00\$00.00~\copyright~2025 IEEE}

\maketitle

\input{sec/0_abstract}


\input{sec/1_intro}

\input{sec/2_related}

\input{sec/3_methods}
\input{sec/4_experiments}

\input{sec/5_conclusion}


\bibliographystyle{IEEEtran}
\bibliography{main}

%


 



\vfill

\end{document}

%% file: sec/0_abstract.tex
\begin{abstract}
We introduce SPFSplatV2, an efficient feed-forward framework for 3D Gaussian splatting from sparse multi-view images, requiring no ground-truth poses during training or inference. The framework employs a shared feature extraction backbone to jointly predict 3D Gaussian primitives and camera poses in a canonical space from unposed inputs. To enable efficient and accurate pose estimation, we introduce a masked attention mechanism for target-view pose prediction and a reprojection loss that enforces pixel-aligned Gaussian primitives, providing stronger geometric constraints.
We further demonstrate the compatibility of our training framework with different reconstruction architectures, resulting in two model variants. Remarkably, despite the absence of pose supervision, our method achieves state-of-the-art performance in both in-domain and out-of-domain novel view synthesis, even under extreme viewpoint changes and limited image overlap. It also surpasses \revision{many} methods that rely on geometric supervision in relative pose estimation. By eliminating dependence on ground-truth poses, our method offers the scalability to leverage larger and more diverse datasets. Code and pretrained models will be available on our project page: \url{https://ranrhuang.github.io/spfsplatv2/}.

\end{abstract}

%% file: sec/1_intro.tex
\section{Introduction}
\IEEEPARstart{R}ecent advancements in 3D reconstruction and novel view synthesis (NVS) have been driven by Neural Radiance Fields (NeRFs)~\cite{mildenhall2021nerf} and 3D Gaussian splatting (3DGS) ~\cite{kerbl20233dgs}.
A standard training pipeline for novel view synthesis reconstructs a 3D scene from input views and optimizes it by aligning rendered novel views with ground-truth images~\cite{chen2021mvsnerf, xu2024murf, charatan2024pixelsplat, chen2024mvsplat, ye2025noposplat, smart2024splatt3r}.

\input{figure/training_pipeline_comparison}
State-of-the-art methods typically employ geometry-aware architectures by constructing cost volumes~\cite{chen2021mvsnerf,xu2024murf,chen2024mvsplat}, leveraging epipolar transformers~\cite{charatan2024pixelsplat}, or encoding camera poses using Pl\"ucker ray embeddings~\cite{zhang2024gslrm, tang2024lgm, xu2024grm}. These approaches rely on camera poses estimated with Structure-from-Motion (SfM)\cite{schonberger2016sfm} to reconstruct 3D scenes, as illustrated in Fig.~\ref{fig:pipeline_comparison} (a). However, acquiring camera poses from SfM is computationally expensive and often unreliable in sparse-view scenarios due to insufficient correspondences, limiting the applicability of these \textit{pose-required} methods. To address this, recent research has focused on novel view synthesis under pose-free settings.

\input{figure/our_pipeline}
Existing pose-free methods reconstruct 3D scenes from unposed images by learning in a canonical space~\cite{jiang2024leap, wang2024pflrm, smart2024splatt3r, ye2025noposplat}, leveraging latent scene representations~\cite{kani2024upfusion, sajjadi2022upsrt}, or jointly optimizing both context-view camera poses and 3D scene representations~\cite{hong2024coponerf, chen2023dbarf, lin2021barf}. Although these methods do not require accurate poses at inference, their training is still supervised by rendering losses given ground-truth poses at novel viewpoints, as shown in Fig.~\ref{fig:pipeline_comparison} (b).
We therefore categorize these approaches as \textit{supervised pose-free} methods.
As a result, their reliance on training datasets with known camera poses restricts the scalability to large-scale real-world data without pose annotations.
\IEEEpubidadjcol 

This raises a critical question: \textit{Are ground-truth poses truly indispensable for optimizing 3D scene representations during training}? One solution is to use estimated poses at novel viewpoints, referred to as the \textit{self-supervised pose-free} paradigm in Fig.~\ref{fig:pipeline_comparison} (c). However, this presents an inherent challenge: since the rendering loss intrinsically couples the learning of 3D scene geometry and camera poses, pose errors can degrade reconstruction quality, which further hampers pose estimation. Such mutual dependency creates a feedback loop that can potentially lead to unstable training or even divergence. Recent self-supervised pose-free approaches~\cite{hong2024pf3plat, kang2025selfsplat} struggle to mitigate this issue primarily due to their use of separate and cascading modules for scene reconstruction and pose estimation, discouraging the learning of consistent feature representations across the two tasks and impairing geometric consistency. 
Consequently, these methods exhibit poor training stability, particularly under large viewpoint changes, and still lag far behind state-of-the-art pose-required and supervised pose-free methods~\cite{chen2024mvsplat, charatan2024pixelsplat, ye2025noposplat}.

To address the challenge, we introduce SPFSplatV2, a \textbf{s}elf-supervised \textbf{p}ose-\textbf{f}ree approach for 3D Gaussian splatting from unposed sparse views. 
As shown in Fig.~\ref{fig:overview}, SPFSplatV2 employs a shared backbone for feature extraction with dedicated heads for predicting 3D Gaussian primitives and camera poses relative to a reference view. 
The unified backbone improves computational efficiency and  facilitates joint feature learning for scene reconstruction and pose estimation, thereby enhancing geometric consistency and mitigating feedback instability. This is achieved by enabling 3D geometry to benefit from context-aware camera alignment and allowing pose predictions to leverage global scene context.

During training, in addition to context images, the target images are also incorporated as input for target pose estimation, enabling rendering losses at target views. To prevent information leakage from target images into the Gaussian reconstruction of context views, we introduce a masked attention mechanism, as shown in Fig.~\ref{fig:overview}. In this design, context tokens attend only to context tokens, ensuring that 3D Gaussian reconstruction remains independent of target-view information. Conversely, target tokens attend to both context and target tokens, allowing the model to exploit global scene context for accurate target pose estimation.
Finally, we complement the rendering loss with a reprojection loss that explicitly enforces alignment between the predicted Gaussians and their corresponding image pixels, imposing stronger geometric constraints and further enhancing training stability.
In conclusion, we make the following key contributions:
\begin{itemize}
    \item We propose SPFSplatV2, a feed-forward framework with masked attention that enables efficient and stable joint optimization of scene reconstruction and pose estimation from sparse unposed views, requiring no ground-truth poses during training and inference.
    \item SPFSplatV2 outperforms state-of-the-art pose-required, supervised pose-free, and self-supervised pose-free methods on both in-domain and out-of-domain novel view synthesis, demonstrating robustness under limited view overlap and extreme viewpoint changes. Despite relying solely on image supervision, its efficient feed-forward relative pose estimation surpasses many approaches that depend on geometric supervision.
    \item 
    By eliminating the reliance on ground-truth poses during training, our method offers the scalability needed to leverage larger and more diverse datasets. Its effectiveness across different architectures further demonstrates the paradigm’s broad compatibility. 
\end{itemize}

This work substantially extends our previous method, SPFSplat~\cite{huang2025spfsplat}, with the key novelties summarized as follows:

\begin{itemize}
\item \textit{Methodological Improvements}: Different from SPFSplat, which employs separate context-only and context-with-target input branches to avoid target information leakage, we introduce a unified architecture with masked attention mechanism that reduces computational overhead and pose misalignment.
\revision{We further adopt learnable pose tokens for pose estimation to ensure compatibility with the VGGT architecture.}
In addition, a multi-view dropout strategy enhances generalization across varying numbers and spatial distributions of context views.

\item \textit{Architectural Compatibility}: We demonstrate that our training paradigm is compatible with state-of-the-art reconstruction models. To this end, we develop two variants: SPFSplatV2, which follows a MASt3R-style~\cite{leroy2024mast3r} architecture (consistent with SPFSplat), and SPFSplatV2-L, which adopts the VGGT~\cite{wang2025vggt} architecture.

\item \textit{Superior Performance}: Extensive experiments show that SPFSplatV2 and SPFSplatV2-L achieve significant improvements over SPFSplat~\cite{huang2025spfsplat} and other state-of-the-art methods in novel view synthesis, cross-domain generalization and relative pose estimation. 

\end{itemize}

%% file: figure/training_pipeline_comparison.tex
\begin{figure}[th]
    \centering
    \includegraphics[width=\linewidth]{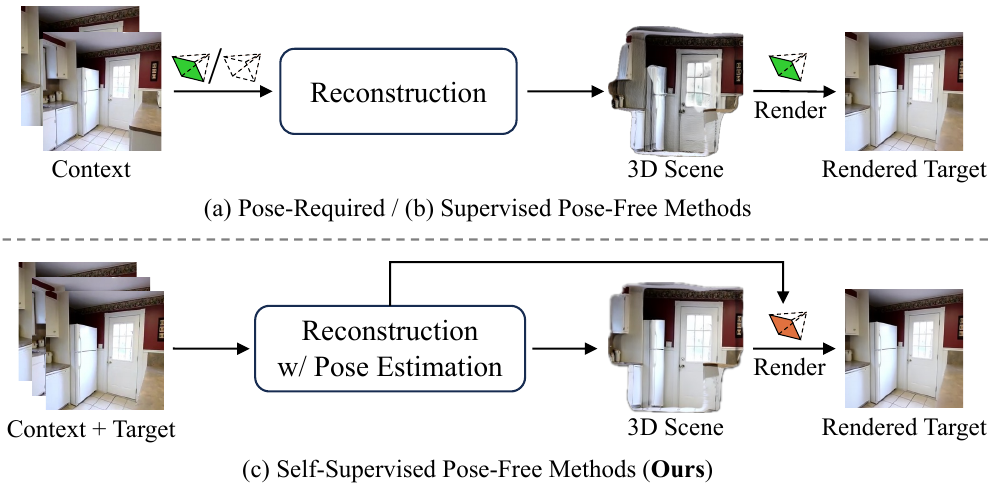}
    \caption{Comparison of three typical \textbf{training} pipelines for sparse-view 3D reconstruction in novel view synthesis. For simplicity, the image rendering loss on the rendered target view is omitted. 
    (a)  Pose-required methods rely on {\color{numbergreen}ground-truth poses} for both 3D scene reconstruction and target-view rendering. (b) Supervised pose-free methods require no ground-truth poses for reconstruction but still rely on {\color{numbergreen}ground-truth poses} for rendering loss. 
    (c)  Our self-supervised pose-free pipeline instead leverages {\color{myorange}estimated target poses} to optimize 3D scene reconstruction from unposed images, thereby removing dependence on ground-truth poses during both training and inference.}
    \label{fig:pipeline_comparison}
\end{figure}

%% file: figure/our_pipeline.tex
\begin{figure*}[ht]
    \centering
    \includegraphics[width=1\textwidth]{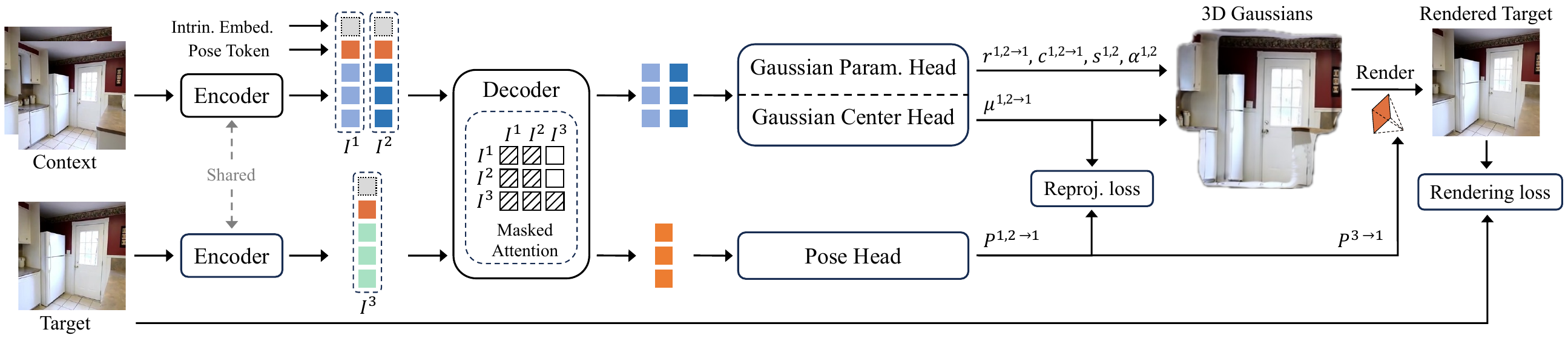}
    \caption{Training pipeline of SPFSplatV2.    A shared backbone with three specialized heads simultaneously predicts Gaussian centers, additional Gaussian parameters, and camera poses from unposed images in a canonical space, with the first input view as the reference. Encoder tokens, concatenated with a learnable pose token and an optional embedding of ground-truth intrinsics, are fed into the decoder, which employs masked attention to prevent context tokens from attending to target tokens, ensuring Gaussian reconstruction remains independent of target-view information. The 3D Gaussians are optimized via a rendering loss using the predicted target poses, while a reprojection loss enforces alignment between Gaussian centers and their corresponding pixels using the predicted context poses. By jointly optimizing Gaussians and camera poses, the pipeline enhances geometric consistency and improves reconstruction quality. }
    \label{fig:overview}
\end{figure*}

%% file: sec/2_related.tex
\section{Related Work}
\label{sec:related}

\subsection{Novel View Synthesis}
NeRF~\cite{mildenhall2021nerf} and 3DGS~\cite{kerbl20233dgs} have demonstrated strong performance in 3D reconstruction and novel view synthesis (NVS). Early methods rely on dense input views for per-scene optimization~\cite{muller2022instant,fridovich2023kplane,chen2022tensorf, fridovich2022plenoxels}, whereas recent approaches enable generalizable reconstruction from sparse-view images~\cite{chen2021mvsnerf, xu2024murf, charatan2024pixelsplat, chen2024mvsplat, tang2024lgm, zhang2024gslrm, xu2024grm, ye2025noposplat, smart2024splatt3r, ye2026yonosplat, xu2025depthsplat, lao2026lgtm}. Typical NVS pipelines reconstruct 3D scenes from input views and optimize them by aligning synthesized images to ground-truth targets. Based on their dependence on ground-truth camera poses during training and inference, existing methods can be grouped into pose-required, supervised pose-free, and self-supervised pose-free approaches, as in Fig.~\ref{fig:pipeline_comparison}.

\textit{Pose-Required Methods} reconstruct 3D scenes from images given accurate poses using geometry-aware architectures~\cite{chen2021mvsnerf, xu2024murf, charatan2024pixelsplat, chen2024mvsplat, tang2024lgm, zhang2024gslrm, xu2024grm, xu2025depthsplat}. For example, MVSNeRF~\cite{chen2021mvsnerf} and MuRF~\cite{xu2024murf} construct cost volumes for multi-view aggregation to reconstruct radiance fields, while MVSplat~\cite{chen2024mvsplat} uses cost volumes for depth estimation to reconstruct Gaussian primitives. Other strategies include epipolar transformers in pixelSplat~\cite{charatan2024pixelsplat} or encoding camera poses with Pl\"ucker ray embeddings~\cite{zhang2024gslrm, tang2024lgm, xu2024grm} \revision{or leveraging pretrained
depth priors~\cite{xu2025depthsplat}}. Despite their effectiveness, these methods depend on SfM for precise camera poses, which is computationally expensive and often unreliable in sparse-view scenarios. Recent pose estimation techniques~\cite{wang2024dust3r, leroy2024mast3r, zhang2022relpose, lin2024relpose++, wang2023posediffusion} mitigate some issues but still struggle in low-overlap or texture-less settings. Consequently, pose-required methods remain impractical for unposed reconstruction during both training and inference.

 \textit{Supervised Pose-Free Methods} enable 3D reconstruction from unposed images, relaxing the need for camera poses at inference. Methods such as UpSRT~\cite{sajjadi2022upsrt} and UpFusion~\cite{kani2024upfusion} encode unposed images into latent scene representations, while BARF~\cite{lin2021barf}, SPARF~\cite{truong2023sparf}, DBARF~\cite{chen2023dbarf}, and CoPoNeRF~\cite{hong2024coponerf} jointly optimize poses and NeRF representations. LEAP~\cite{jiang2024leap} and PF-LRM~\cite{wang2024pflrm} leverage ViT architectures to define neural volumes in canonical camera coordinates. More recently, Splatt3R~\cite{smart2024splatt3r} predicts 3D Gaussians in a canonical space by regressing offsets to pointmaps from a frozen MASt3R~\cite{leroy2024mast3r}, but requires depth supervision. NoPoSplat~\cite{ye2025noposplat} removes this depth reliance and refines this pipeline by fine-tuning MASt3R and incorporating intrinsics to mitigate scale ambiguity. \revision{YoNoSplat~\cite{ye2026yonosplat} provides a flexible architecture that supports both pose-required and pose-free input settings.} However, despite removing pose requirements at inference, these methods still depend on ground-truth poses during training via rendering losses~\cite{jiang2024leap, wang2024pflrm, smart2024splatt3r, ye2025noposplat, sajjadi2022upsrt, kani2024upfusion, xu2025depthsplat, ye2026yonosplat}, explicit pose supervision~\cite{hong2024coponerf, ye2026yonosplat}, or coarse pose initialization~\cite{lin2021barf, truong2023sparf}, therefore limiting scalability to large-scale unposed data.

 \textit{Self-Supervised Pose-Free Methods} completely eliminate the reliance on ground-truth poses during training by enabling rendering losses at novel viewpoints using estimated poses, as shown in Fig.~\ref{fig:pipeline_comparison} (c).
For instance, Nope-NeRF~\cite{bian2023nope}, CF-3DGS~\cite{fu2024colmap}, and FlowCam~\cite{smith2023flowcam} reconstruct 3D scenes and estimate camera poses incrementally by re-rendering dense video sequences. However, they are limited to continuous video frames and do not generalize well to sparse views.
Recent self-supervised pose-free methods, such as PF3plat~\cite{hong2024pf3plat} and SelfSplat~\cite{kang2025selfsplat}, attempt to estimate both input- and target-view poses from sparse views. PF3plat relies on off-the-shelf feature descriptors~\cite{lindenberger2023lightglue} together with RANSAC-based initialization, resulting in a pipeline that is computationally expensive and not end-to-end trainable. In contrast, SelfSplat employs cross-view U-Nets~\cite{ronneberger2015unet,rombach2022diffusion} for pose prediction, but its performance degrades substantially under large viewpoint changes.
Beyond the limitations of pose estimation, both methods decouple pose prediction and Gaussian reconstruction into separate modules, preventing feature sharing between the two tasks and leading to weaker geometric alignment and higher computational overhead. 
\revision{AnySplat~\cite{jiang2025anysplat} also jointly optimizes 3D Gaussians and camera poses. However, its rendering loss is applied only to the input context views, providing no direct supervision for novel viewpoints during training. As a result, novel view synthesis performance remains limited, and the method relies on continuous distillation from VGGT throughout training to stabilize depth and camera pose estimation.}

SPFSplat~\cite{huang2025spfsplat}, our previous approach, also adopts a self-supervised, pose-free paradigm by jointly optimizing 3D Gaussians and camera poses through a shared backbone in a canonical space. Guided by image rendering and reprojection losses, the shared representation promotes geometric consistency and stable optimization. \revision{Unlike AnySplat, SPFSplat uses self-predicted target poses to render and supervise target-view observations, providing direct supervision for novel viewpoints. Building upon SPFSplat, we introduce masked attention and additional architectural refinements, resulting in a unified self-supervised framework that is compatible with different backbone designs while improving performance and generalization across varying numbers of input views.}


\subsection{Structure-from-Motion (SfM)}
Structure-from-Motion (SfM)\cite{hartley2003multiple, schonberger2016structure} is a core problem in computer vision that jointly estimates camera parameters and reconstructs sparse 3D maps from image collections. Classical SfM pipelines typically involve local feature detection and matching\cite{lowe2004distinctive, bay2008speeded, rublee2011orb}, geometric verification via epipolar geometry or homographies with RANSAC~\cite{fischler1981ransac}, triangulation~\cite{hartley1997triangulation} to recover 3D points, and bundle adjustment~\cite{triggs1999bundle} to refine poses and structure.
Recent advances have incorporated learning-based components into SfM, including robust feature descriptors~\cite{detone2018superpoint, Dusmanu2019d2,huang2023drkf,luo2020aslfeat}, improved image matching~\cite{sarlin2020superglue, lindenberger2023lightglue}, detector-free matching~\cite{sun2021loftr}, and neural bundle adjustment~\cite{tang2018banet,lin2021barf}. However, the sequential design of SfM pipelines remains prone to error propagation. To overcome this, fully differentiable pipelines have been introduced~\cite{wang2024vggsfm, wang2024dust3r, leroy2024mast3r, tang2025mvdust3r, yang2025fast3r, wang2025vggt}. For example, VGGSfM~\cite{wang2024vggsfm} enables end-to-end sparse reconstruction, while DUSt3R~\cite{wang2024dust3r}, following  the architecture of CroCo~\cite{weinzaepfel2022croco}, performs dense 3D reconstruction without camera parameters. MASt3R~\cite{leroy2024mast3r} further enhances feature matching and local representations but, like DUSt3R, remains constrained by pairwise architectures and costly global optimization, \revision{limiting its applicability to efficient multi-view reconstruction}. Extensions such as MV-DUSt3R+\cite{tang2025mvdust3r} and Fast3R\cite{yang2025fast3r} address multi-view reconstruction, while FLARE~\cite{zhang2025flare} employs cascaded learning with pose as the central bridge. Recently, VGGT~\cite{wang2025vggt} introduces a feed-forward transformer that jointly infers camera parameters, depth, point maps, and 3D tracks, achieving state-of-the-art results.

Similar to these SfM methods, our method jointly predicts 3D points and poses, with rendering and reprojection losses acting as a differentiable form of bundle adjustment to refine both geometry and poses. Unlike prior work, it requires no ground-truth geometric priors during training. In addition, the training paradigm is naturally compatible with advanced reconstruction backbones such as MASt3R~\cite{leroy2024mast3r} and VGGT~\cite{wang2025vggt}, giving rise to two variants: SPFSplatV2 and SPFSplatV2-L.

%% file: sec/3_methods.tex
\section{Method}
\label{sec:method}

We aim to learn a feed-forward network that reconstructs 3D Gaussians from unposed images while simultaneously estimating camera poses. During training, the predicted poses are used to render target views for photometric supervision, enabling joint optimization of geometry and camera poses without requiring ground-truth pose annotations.

\subsection{Problem Formulation}
Consider $N$ context images $\{\boldsymbol{I}^v\}_{v=1}^{N}$ as input. During training, additional $M$ target images $\{\boldsymbol{I}^v\}_{v=N{+}1}^{N{+}M}$ are provided, resulting in a total of $V = N{+}M$ views.


\textit{3D Gaussian Reconstruction:}
Following~\cite{smart2024splatt3r, ye2025noposplat}, we predict 3D Gaussians from context images in a canonical 3D space where the first input view $\boldsymbol{I}^1$ serves as the global  coordinate frame. The reconstruction network is formulated as:
\begin{equation}
    f_{\boldsymbol{\theta}} : \{\boldsymbol{I}^v\}_{v=1}^{N} \mapsto \{   \boldsymbol{\mathcal{G}}^{v \rightarrow 1}\}_{v=1}^{N}, 
\label{eq:gaussian_formulation}
\end{equation}
where  $\boldsymbol{\mathcal{G}}^{v \rightarrow 1} = \{(\boldsymbol{\mu}_j^{v \rightarrow 1},  \boldsymbol{r}_j^{v \rightarrow 1},  \boldsymbol{s}_j^v, \boldsymbol{c}_j^{v \rightarrow 1}, \alpha_j^v)\}_{\substack{j=1,\dots,H \times W}}$ represents the pixel-aligned Gaussians for $\boldsymbol{I}^v$, represented in the coordinate frame of $\boldsymbol{I}^1$. 
Each Gaussian is parameterized by center $ \boldsymbol{\mu} \in \mathbb{R}^3 $, rotation quaternion $\boldsymbol{r} \in \mathbb{R}^4 $, scale $\boldsymbol{s} \in \mathbb{R}^3 $, opacity $ \alpha \in \mathbb{R} $, and spherical harmonics (SH) $ \boldsymbol{c} \in \mathbb{R}^k $ with \( k \) degrees of freedom. 

\textit{Pose Estimation:}
We introduce a pose network $f_{\boldsymbol{\phi}}$ to estimate the relative transformation from each view $\boldsymbol{I}^v$ to the reference view $\boldsymbol{I}^1$, which is denoted as $\boldsymbol{P}^{v \to 1} =[\boldsymbol{R}^{v \to 1} | \boldsymbol{T}^{v \to 1} ]$, where $\boldsymbol{R}^{v \to 1} \in \mathbb{R}^{3 \times 3}$ represents the rotation matrix, and $\boldsymbol{T}^{v \to 1} \in \mathbb{R}^{3 \times 1}$ represents the translation vector.  This can be formulated as:
\begin{equation}
    \boldsymbol{P}^{v \to 1} = f_{\boldsymbol{\phi}}(\boldsymbol{I}^{v},\dots, \boldsymbol{I}^{1}), v \in [1, \dots, V].
\label{eq:pose_head}
\end{equation}

\textit{Novel View Synthesis:}
Novel views are then rendered using the estimated target poses and reconstructed Gaussians:
\begin{equation}
    \begin{aligned}
        \hat{\boldsymbol{I}^t} &= \mathcal{R}(\boldsymbol{P}^{t \to 1}, \{\boldsymbol{\mathcal{G}}^{v \rightarrow 1}\}_{v=1}^{N}), & t \in [N{+}1, \dots, V].
    \end{aligned}
\label{eq:image_rendering}
\end{equation}

\subsection{Architecture}
Following state-of-the-art large reconstruction models such as MASt3R~\cite{leroy2024mast3r} and VGGT~\cite{wang2025vggt}, our framework consists of three main components: an encoder, a decoder, and task-specific prediction heads, as illustrated in Fig.~\ref{fig:overview}. Both the encoder and decoder follow Vision Transformer (ViT) architectures~\cite{dosovitskiy2021vit}.
As shown in Fig.~\ref{fig:detailed_arch}, we develop two model variants: SPFSplatV2, which adopts the MASt3R-style architecture, and SPFSplatV2-L, which follows the VGGT design. In the following, we introduce both variants in detail.

\textit{Encoder:}
The RGB image $\boldsymbol{I}^v$ for each input view $v$ is first patchified and flattened into a sequence of image tokens. These tokens are independently processed by a shared-weight ViT encoder, which extracts view-specific feature representations $\boldsymbol{F}^{v} \in \mathbb{R}^{L \times C}$, where $L$ denotes the number of tokens. This is formulated as follows:
\begin{equation}
    \boldsymbol{F}^{v} = \mathrm{Encoder}(\boldsymbol{I}^{v}), v \in [1, \dots, V].
\label{eq:encoder}
\end{equation}


\input{figure/detailed_arch}

\textit{Learnable Pose Token:}
MASt3R does not provide a dedicated pose estimation head, and the original SPFSplat~\cite{huang2025spfsplat} estimates camera poses through global average pooling of the decoded image tokens. \revision{To maintain compatibility with VGGT and provide a unified pose estimation design across architectures, we adopt learnable pose tokens in both SPFSplatV2 and SPFSplatV2-L.} SPFSplatV2 introduces a learnable pose token $\boldsymbol{g} \in \mathbb{R}^{1 \times C}$, which is replicated for each view $v \in [1,V]$ as $\boldsymbol{g}^v$. 
Unlike SPFSplatV2 which uses an asymmetrical decoder  to distinguish the reference frame from the other views, 
SPFSplatV2-L introduces two separate learnable pose tokens $\bar{\boldsymbol{g}}$ and $\bar{\bar{\boldsymbol{g}}} \in \mathbb{R}^{1 \times C}$, following VGGT. Specifically, $\bar{\boldsymbol{g}}$ is assigned to the reference frame ($\boldsymbol{g}^1 := \bar{\boldsymbol{g}}$), while all other views share $\bar{\bar{\boldsymbol{g}}}$ ($\boldsymbol{g}^v := \bar{\bar{\boldsymbol{g}}}, v \in [2, \dots, V]$). The pose tokens are concatenated with the encoder tokens, yielding $\boldsymbol{F}^{v} := [\boldsymbol{g}^v, \boldsymbol{F}^{v}]$. In the decoding stage, the learnable pose tokens can selectively attend to the informative features.

\textit{Intrinsics Embedding:}
Following~\cite{ye2025noposplat}, we encode the camera intrinsics of each view into a token $\boldsymbol{k}^v$ via a linear layer and concatenate it with the pose token and encoder tokens, forming the decoder input $\boldsymbol{F}^v := [\boldsymbol{k}^v, \boldsymbol{g}^v, \boldsymbol{F}^v]$. This explicitly injects calibration information, helping to resolve scale ambiguity and improve alignment of predicted poses and 3D Gaussians, particularly under large focal length variations. Importantly, the intrinsic token is optional, and SPFSplatV2 maintains strong performance without it (Sec.~\ref{sec:experiments_ablation}), underscoring the robustness and flexibility of the design.

\textit{Masked Multi-view Decoder}: 
To effectively aggregate information across multiple views, we adopt a ViT-based decoder with cross-view attention, enabling joint reasoning over token representations across input views and facilitating cross-view information exchange to capture spatial
relationships and the global 3D scene geometry. 

\input{figure/masked_attention}

During training, since target views are also provided as input, the original SPFSplat~\cite{huang2025spfsplat} adopts a dual-branch design to avoid information leakage from target views into the Gaussian reconstruction of context views. One branch processes only context images for Gaussian prediction, while the other takes both context and target images for pose estimation. However, this design introduces two drawbacks: (i) higher computational cost in cross-attention caused by two forward passes during training, as shown in Fig.~\ref{fig:attention_mask} (a), and (ii) target poses and 3D geometry are predicted in separate branches, which may lead to pose-geometry misalignment.

To address these issues, we introduce a \textit{masked attention} mechanism as shown in Fig.~\ref{fig:detailed_arch} and Fig.~\ref{fig:attention_mask} (b). In this approach, context and target images are jointly processed in a single forward pass, with cross-attention selectively masked to control information flow. Specifically, context tokens attend only to context tokens, ensuring Gaussian reconstruction remains independent of target-view information. Meanwhile, target tokens attend to both context and target tokens, allowing the network to leverage global cues for accurate pose estimation.

For SPFSplatV2, we extend MASt3R’s pairwise asymmetric decoder to a multi-view setting, which scales efficiently with the number of views while avoiding excessive memory overhead, following similar implementations in~\cite{ye2025noposplat, tang2025mvdust3r}. Each decoder block first performs intra-view self-attention, followed by masked cross-attention. Tokens from the first (reference) view are processed with $\mathrm{DecoderBlock}^1$, while tokens from the remaining views use $\mathrm{DecoderBlock}^2$. The two decoders share the same architecture but maintain independent weights. Formally, the masked decoder block is defined as:
\begin{equation}
\small 
\boldsymbol{G}^{v}_{i} =
\begin{cases}


\mathrm{DecoderBlock}_i^1(\boldsymbol{G}^{v}_{i-1}, \boldsymbol{G}^{1:K}_{i-1}), & v=1, \\[3pt]
\mathrm{DecoderBlock}_i^2(\boldsymbol{G}^{v}_{i-1}, \boldsymbol{G}^{1:K}_{i-1}), & v \in [2, \dots, V], \\
\end{cases}
\label{eq:v2_decoder}
\end{equation}
for $i = 1, \dots, B$, where $B$ is the number of decoder blocks and $\boldsymbol{G}^{v}_{0}=\boldsymbol{F}^v$ are the initial tokens for view $v$.
For context views ($v \in [1,\dots,N]$), we set $K=N$ such that attention is restricted to the context views. For target views ($v \in [N+1,\dots,V]$), we set $K=V$, allowing attention over all views.


For SPFSplatV2-L, we adopt the VGGT architecture, which alternates between intra-frame and inter-frame attention. 
Different from MASt3R’s asymmetric design, the decoder here is unified across all views,
which can be expressed as:

\begin{equation}
\small
\boldsymbol{G}^{v}_{i} = \mathrm{DecoderBlock}_i\left(\boldsymbol{G}^{v}_{i-1}, \boldsymbol{G}^{1:K}_{i-1}\right), \quad v \in [1, \dots, V],
\label{eq:v2l_decoder}
\end{equation}
for $i = 1, \dots, B$, where $B$ is the number of decoder blocks, and $K$ follows the same definition as in SPFSplatV2.


Overall, the masked multi-view decoder achieves three key advantages: (i) it preserves generalization to novel viewpoints by strictly preventing target-specific information from contaminating the Gaussian representation, (ii) it significantly reduces computational overhead by avoiding redundant forward passes, and (iii) it predicts 3D geometry and target poses from a unified decoder representation, thereby improving pose-geometry alignment and reducing pose drift. Together, these improvements lead to more efficient and stable training, as well as more accurate reconstructions. 

\textit{Gaussian Prediction Heads:}
Following~\cite{smart2024splatt3r, ye2025noposplat}, we employ two DPT-based heads~\cite{ranftl2021dpt} to infer Gaussian parameters.
The first head processes decoder tokens of context views and predicts 3D coordinates for each pixel, defining Gaussian centers. The second head estimates rotation, scale, opacity, and SH coefficients for each Gaussian primitive. 

For SPFSplatV2, the Gaussian center head extends MASt3R’s pairwise asymmetric pointmap head to a multi-view setting by assigning decoder tokens from the first view to the reference head $\mathrm{PointHead}^1$, and tokens from all remaining views to the non-reference head $\mathrm{PointHead}^2$. The Gaussian parameter head follows the same structure as the Gaussian center head. As proposed in~\cite{charatan2024pixelsplat, chen2024mvsplat, ye2025noposplat}, we incorporate high-resolution skip connections by feeding the original context images into the Gaussian parameter heads, preserving fine-grained spatial details. These heads can be formulated as:

\begin{align}
\boldsymbol{\mu}^{ v \rightarrow 1} &=
\begin{cases}
\mathrm{PointHead}^1\!\left(\{\boldsymbol{G}^{v}_{i}\}_{i=0}^{B}\right), & v = 1, \\
\mathrm{PointHead}^2\!\left(\{\boldsymbol{G}^{v}_{i}\}_{i=0}^{B}\right), & v \in [2, \dots, N],
\end{cases}
\label{eq:v2_pts3d_head} \\[10pt]
\boldsymbol{\overline{\mathcal{G}}}^{v \rightarrow 1} &=
\begin{cases}
\mathrm{GSHead}^1\!\left(\{\boldsymbol{G}^{v}_{i}\}_{i=0}^{B}, \boldsymbol{I}^{v}\right), & v = 1,   \\
\mathrm{GSHead}^2\!\left(\{\boldsymbol{G}^{v}_{i}\}_{i=0}^{B}, \boldsymbol{I}^{v}\right), &  v \in [2, \dots, N],
\end{cases}
\label{eq:v2_gs_head}
\end{align}
where $\{\boldsymbol{G}^{v}_{i}\}_{i=0}^{B}$  denotes the set of decoder tokens taken from different blocks,  $\boldsymbol{\mu}^{ v \rightarrow 1}$ denotes Gaussian centers, and $\boldsymbol{\overline{\mathcal{G}}}^{v \rightarrow 1}= \{(\boldsymbol{r}_j^{v \rightarrow 1},  \boldsymbol{c}_j^{v \rightarrow 1}, \alpha_j^v, \boldsymbol{s}_j^v)\}$ represents rotation, scale, opacity, and SH coefficients for each Gaussian primitive.

For SPFSplatV2-L, we adopt the VGGT design for the pointmap head, which serves as both the Gaussian center head and the Gaussian parameter head. Similar to SPFSplatV2, the Gaussian parameter head also incorporates the original context images as an auxiliary input. Unlike the asymmetric design in SPFSplatV2, the Gaussian prediction heads in SPFSplatV2-L are unified across all views:
\begin{align}
    \boldsymbol{\mu}^{ v \rightarrow 1} &= \mathrm{PointHead}\!\left(\{\boldsymbol{G}^{v}_{i}\}_{i=0}^{B}\right), &  v \in [1, \dots, N]\\
    \boldsymbol{\overline{\mathcal{G}}}^{v \rightarrow 1}  &= \mathrm{GSHead}\!\left(\{\boldsymbol{G}^{v}_{i}\}_{i=0}^{B}, \boldsymbol{I}^{v}\right), &  v \in [1, \dots, N].
\label{eq:v2l_gs_head}
\end{align}


\textit{Pose Head:}
For SPFSplatV2, the attended pose tokens $\hat{\boldsymbol{g}^v}$ are fed into the pose head and further processed by a 3-layer MLP to predict the camera pose as a 10-dimensional representation~\cite{brachmann2023accelerated}. The predicted pose representation is decomposed into translation and rotation for each view. The translation is represented using four homogeneous coordinates~\cite{brachmann2023accelerated}, 
while the rotation is encoded in a 6D format, capturing two unnormalized coordinate axes. These axes are normalized and combined via a cross-product operation to construct a full rotation matrix~\cite{zhou2019continuity}.
To compute the relative pose with respect to the reference view, the 10D pose representation is converted into a homogeneous transformation matrix $\boldsymbol{P}^{v \to 1} \in \mathbb{R}^{4 \times 4}$.
Following MASt3R, we make the pose head asymmetrical:
\begin{equation}
\boldsymbol{P}^{ v \rightarrow 1} =
\begin{cases}
\mathrm{PoseHead}^1(\hat{\boldsymbol{g}^v}), & v = 1, \\
\mathrm{PoseHead}^2(\hat{\boldsymbol{g}^v}), &  v \in [2, ..., V],
\end{cases}
\label{eq:v2_pose_head}
\end{equation}
where $\boldsymbol{P}^{v \to 1}$ is the estimated relative pose from $\boldsymbol{I}^v$  to $\boldsymbol{I}^1$.

For SPFSplatV2-L, the pose head follows the original VGGT design: the refined pose tokens $\hat{\boldsymbol{g}}^v$ are subsequently processed by four additional self-attention layers and a linear projection to predict the camera parameters.
\begin{equation}
\boldsymbol{P}^{ v \rightarrow 1} = \mathrm{PoseHead}(\hat{\boldsymbol{g}^v}), v \in [1,..., V].
\label{eq:v2l_pose_head}
\end{equation}

We normalize the predicted poses relative to the first view, such that the first view has the canonical pose $[\mathbf{U} \mid \mathbf{0}]$, where $\mathbf{U}$ is the identity matrix and $\mathbf{0}$ is the zero translation vector.

\subsection{Loss Functions}
\textit{Image Rendering Loss:}
Our model is trained using ground-truth target RGB images as supervision. The training loss is defined as a weighted combination of the $L_2$ loss and the LPIPS loss~\cite{zhang2018lpips}:
\begin{equation}
    \mathcal{L}_{\text{render}} = \| \boldsymbol{I}^t - \hat{\boldsymbol{I}^t}\|_2 + \gamma \text{LPIPS}(\boldsymbol{I}^t, \hat{\boldsymbol{I}^t}),
\label{eq:rendering_loss}
\end{equation}
where $\boldsymbol{I}^t$ and $\hat{\boldsymbol{I}}^t$ denote the ground-truth and rendered target images for $t \in [N+1, V]$, and $\gamma$ is a weighting factor that balances pixel-level accuracy and perceptual similarity.

\textit{Reprojection Loss:}
Existing approaches enforce pixel-aligned Gaussian prediction by constraining Gaussian locations along the input viewing rays~\cite{charatan2024pixelsplat, chen2024mvsplat, xu2024grm, zhang2024gslrm, hong2024pf3plat, kang2025selfsplat}. Meanwhile, canonical-space-based methods~\cite{smart2024splatt3r, ye2025noposplat} rely on ground-truth camera poses to guide the canonical 3D points (Gaussian centers).
Both strategies ensure alignment between each pixel and its corresponding 3D point. However, since our model learns 3D Gaussian centers in a canonical space without known camera poses, the network lacks explicit geometric constraints to enforce pixel-aligned Gaussian representation.

A naive solution is to include context views in the image rendering loss (Eq. \ref{eq:rendering_loss}) by synthesizing images from them and computing the loss against their ground-truth counterparts. However, this leads to unstable training due to overfitting. Specifically, the network prioritizes improving the rendering quality of the first context view, as the 3D Gaussian space is defined in its camera coordinate, making its rendering independent of the learnable poses. 
Since the Gaussians from this view already capture sufficient scene information, the model suppresses the contribution of other context views by shifting their Gaussian centers away and adjusting camera poses, ultimately causing training collapse.

To address this issue, we instead employ a pixel-wise reprojection loss to jointly optimize 3D points and camera poses~\cite{brachmann2024scr, schonberger2016sfm}. Unlike purely image-based supervision, this reprojection loss enforces explicit geometric constraints, thereby reducing overfitting to the context views. Concretely, for each pixel $\mathbf{p}^v_j$ in view $v \in [1, N]$, we project the corresponding 3D Gaussian center $\boldsymbol{\mu}_{j}^{v \rightarrow 1}$ from the canonical coordinate frame into the 2D pixel space using the estimated pose of view $v$, and minimize the reprojection error:
\begin{align}
\mathcal{L}_{\text{reproj}} &= \sum_{v=1}^{N} \sum_{j=1}^{H\times W}
\left| \mathbf{p}^v_{j} - \pi(\boldsymbol{K}^v, \boldsymbol{P}^{v \rightarrow 1}, \boldsymbol{\mu}_{j}^{v \rightarrow 1}) \right|,
\end{align}
where $\pi$ denotes the camera projection function, $\boldsymbol{K}^v$ the camera intrinsics of view $v$, and $\boldsymbol{P}^{v \rightarrow 1}$ the relative pose from view $v$ to the canonical frame.

Different from SPFSplat, which applies reprojection loss to both context-only and context-with-target branches, SPFSplatV2 predicts a single set of context poses, avoiding redundant supervision and potential pose misalignment. This streamlined design leverages reprojection loss to enable more stable training and efficient optimization of pixel-aligned 3D Gaussians, without requiring ground-truth camera poses.

\subsection{Multi-View Dropout}
Unlike SPFSplat, which trains separate models for different numbers of context views, we use a single unified model. To improve generalization, we introduce a multi-view dropout strategy: for more than two context views, the leftmost and rightmost views are retained while intermediate views are randomly dropped during training. This encourages the network to handle flexible input configurations, improves robustness to varying numbers and spatial distributions of views at test time.



%% file: figure/detailed_arch.tex
\begin{figure}[th]
    \centering
    \includegraphics[width=\linewidth]{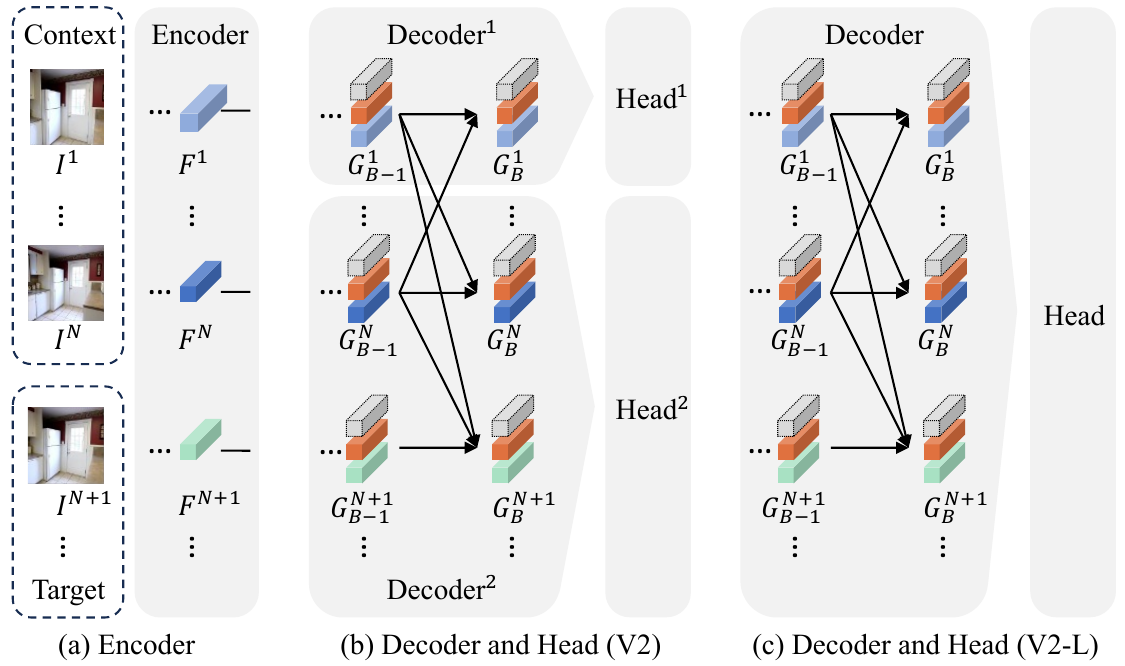}
    \caption{Architecture comparison of SPFSplatV2 and SPFSplatV2-L. SPFSplatV2 (a {+} b) uses asymmetrical decoders and heads to distinguish the reference view $\boldsymbol{I}^1$ from other views, whereas SPFSplatV2-L (a {+} c) employs a unified decoder and head for all views.
    }
    \label{fig:detailed_arch}
\end{figure}

%% file: figure/masked_attention.tex
\begin{figure}[th]
    \centering
    \includegraphics[width=\linewidth]{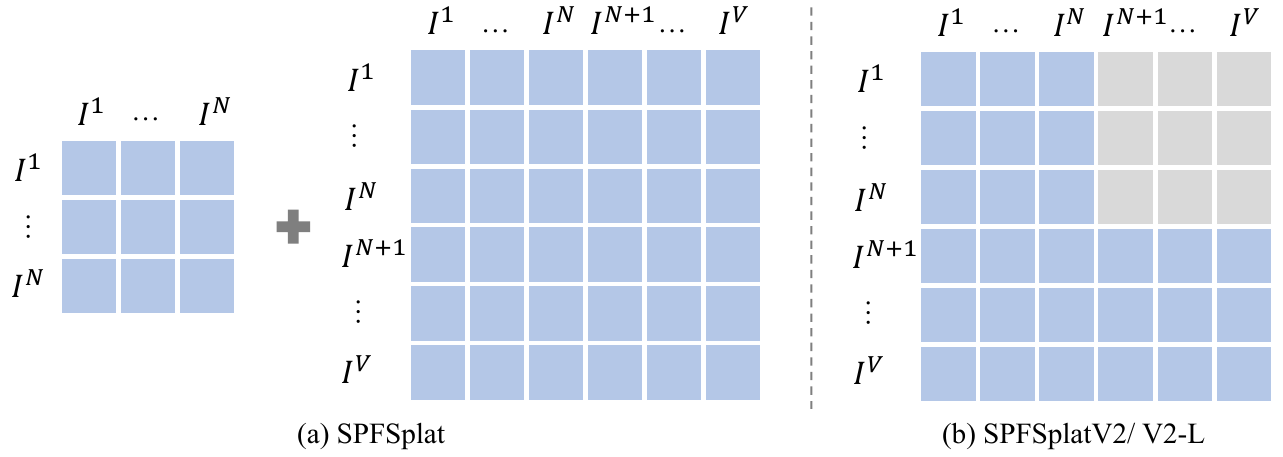}
    \caption{Comparison of cross-attention in (a) SPFSplat and (b) SPFSplatV2/ V2-L.
$I^1, \cdots, I^N$ denote context images, while $I^{N+1}, \cdots, I^V$ denote target images.
SPFSplat relies on dual input branches (context-only and context-with-target) to block target leakage into context reconstruction, leading to higher cross-attention cost.
SPFSplatV2 replaces this with masked attention, enforcing the same separation within a single branch while reducing computation by removing redundant context–target interactions. }
    \label{fig:attention_mask}
\end{figure}


%% file: sec/4_experiments.tex
\section{Experiments}
\label{sec:experiments}

We first introduce the experimental settings and implementation details in Sec.~\ref{sec:experimental_settings} and Sec.~\ref{sec:implementation_details}. We then evaluate the proposed framework under random initialization in Sec.~\ref{sec:ssl_from_scratch}. Subsequently, Sec. \ref{sec:results_with_priors} presents results obtained with pretrained initialization. Finally, comprehensive ablation studies are conducted in Sec.~\ref{sec:experiments_ablation} to analyze the contribution of each component of the proposed framework.

\subsection{Experimental Settings}
\label{sec:experimental_settings}
\textit{Datasets:} In Sec.~\ref{sec:ssl_from_scratch}, we train our method on RealEstate10K (RE10K)~\cite{zhou2018stereo} and evaluate it on both RE10K and out-of-domain ACID~\cite{liu2021infinite}, following the official train-test splits used in prior work~\cite{chen2024mvsplat,charatan2024pixelsplat,ye2025noposplat}. Following~\cite{ye2025noposplat}, evaluations on RE10K and ACID are conducted under varying camera overlaps, where image pairs are grouped into small (0.05--0.3), medium (0.3--0.55), and large (0.55--0.8) overlap ranges using overlap estimates from the pretrained dense matcher RoMa~\cite{edstedt2024roma}.
In Sec.~\ref{sec:results_with_priors}, we train and evaluate our method on both RE10K and ACID for in-domain evaluation. We further evaluate cross-dataset generalization on ACID, DTU~\cite{jensen2014dtu}, DL3DV~\cite{ling2024dl3dv}, and ScanNet++~\cite{yeshwanth2023scannetpp}. \revision{To study scalability, we expand the training data beyond RE10K to include DL3DV, CO3Dv2~\cite{reizenstein2021co3d}, and WildRGBD~\cite{xia2024wildrgbd}.}

\textit{Baselines:}
For novel view synthesis, we compare to three groups of baselines: \textit{pose-required methods} (pixelSplat~\cite{charatan2024pixelsplat}, MVSplat~\cite{chen2024mvsplat}, \revision{DepthSplat~\cite{xu2025depthsplat}, YoNoSplat~\cite{ye2026yonosplat}}), \textit{supervised pose-free methods} (CoPoNeRF~\cite{hong2024coponerf}, Splatt3R~\cite{smart2024splatt3r}, NoPoSplat~\cite{ye2025noposplat}), and \textit{self-supervised pose-free methods} (PF3plat~\cite{hong2024pf3plat}, SelfSplat~\cite{kang2025selfsplat}, SPFSplat~\cite{huang2025spfsplat}, \revision{and AnySplat~\cite{jiang2025anysplat}}). For pose estimation, we compare against \textit{SfM-based methods} (SuperPoint~\cite{detone2018superpoint} + SuperGlue~\cite{sarlin2020superglue}, DUSt3R~\cite{wang2024dust3r}, MASt3R~\cite{leroy2024mast3r}, VGGT~\cite{wang2025vggt}) and \textit{NVS-based methods} (NoPoSplat, SelfSplat, PF3plat, SPFSplat and AnySplat).

\input{table/nvs_from_scratch}

\textit{Evaluation Protocol:} For novel view synthesis, we adopt standard metrics: pixel-level PSNR, patch-level SSIM~\cite{wang2004ssim}, and feature-level LPIPS~\cite{zhang2018lpips}. For pose estimation, following prior works~\cite{sarlin2020superglue, ye2025noposplat}, we report the area under the cumulative pose error curve (AUC) at thresholds of 5$^\circ$, 10$^\circ$, and 20$^\circ$, where the pose error is defined as the maximum of the angular errors in rotation and translation.

During evaluation of novel view synthesis, target images are typically rendered with ground-truth poses~\cite{chen2024mvsplat, charatan2024pixelsplat, hong2024coponerf, smart2024splatt3r}. An alternative is to render using estimated target poses, as in PF3plat~\cite{hong2024pf3plat} and SelfSplat~\cite{kang2025selfsplat}. NoPoSplat~\cite{ye2025noposplat} instead adopts an evaluation-time pose alignment (EPA) strategy, which optimizes the target pose during evaluation while keeping the reconstructed Gaussians fixed, so that the rendered image best matches the ground truth. This alignment decouples rendering quality from pose estimation accuracy, enabling direct assessment of Gaussian reconstruction. In contrast, rendering with estimated poses jointly evaluates reconstruction fidelity and the consistency between estimated poses and the learned Gaussians. Unless otherwise noted, we render with estimated poses for comprehensive evaluation and additionally report results with pose alignment for fair comparison to NoPoSplat.

\input{table/novel_view_synthesis_comparison_on_RE10k}

\input{table/novel_view_synthesis_comparison_on_ACID}
\input{figure/qualitative_comparison_on_RE10k_and_ACID}

\subsection{Implementation Details.}
\label{sec:implementation_details}
Our method is implemented in PyTorch and leverages a CUDA-based 3DGS renderer with gradient support for camera poses. All models are trained on a single NVIDIA A100 GPU. Each training sample corresponds to a scene with context and target views, with the frame distance between context views gradually increased during training. The initial learning rate is set to $1 \times 10^{-5}$ for the backbone and $1 \times 10^{-4}$ for all other parameters, and LPIPS and reprojection losses are weighted 0.05 and 0.001, respectively.
For SPFSplatV2, the encoder adopts a ViT-Large architecture with a patch size of 16, while the decoder is based on ViT-Base. For SPFSplatV2-L, the encoder is a ViT-Large from DINOv2~\cite{oquab2023dinov2} with a patch size of 14.
Training is performed at a resolution of 256 $\times$ 256 and 224 $\times$ 224 for V2 and V2-L, respectively.

\subsection{\revision{Self-Supervised Pose-Free Learning from Scratch}}
\label{sec:ssl_from_scratch}

\revision{
Without geometric constraints or suitable initialization, the Gaussian center head predicts arbitrary 3D coordinates, providing no guarantee of a valid canonical structure and resulting in unstable or noisy gradients during early training. To stabilize optimization from random initialization, NoPoSplat and SPFSplat adopt a short warm-up phase with a point-cloud distillation loss from DUSt3R during early training. In contrast, we find that stable convergence can be achieved without such distillation by imposing a simple canonical constraint on the Gaussian center parameterization. Specifically, we derive a scalar depth value for each pixel by averaging its predicted 3D coordinates, and reproject it into 3D using the predicted camera poses and ground-truth intrinsics. The resulting aligned points serve as Gaussian centers throughout training, enforcing canonical consistency and enabling stable optimization from random initialization.
}

\revision{
Empirically, as shown in Tab.~\ref{tab:nvs-from-scratch}, we make the following observations. 1) When trained from random initialization, NoPoSplat, despite benefiting from both DUSt3R distillation warm-up and ground-truth target pose supervision, fails to produce reliable pose estimates. In contrast, SPFSplatV2 optimizes geometry with self-predicted camera poses and achieves substantially stronger pose estimation performance, highlighting the effectiveness of self-supervised geometric optimization for learning camera poses from scratch. 2) Despite using no pretrained initialization, SPFSplatV2 and SPFSplatV2-L achieve significantly better novel view synthesis performance than existing pose-free methods that rely on pretrained geometric components, including SelfSplat, PF3plat, and AnySplat. This comparison is particularly noteworthy for AnySplat, which is trained on substantially larger mixed datasets and further benefits from VGGT-generated pseudo-label distillation throughout training. While AnySplat achieves higher pose estimation accuracy under these additional advantages, SPFSplatV2 and SPFSplatV2-L obtain superior NVS quality while maintaining competitive pose estimation performance using purely self-supervised optimization. 3) Furthermore, SPFSplatV2 and SPFSplatV2-L achieve comparable performance under random initialization despite their different architectural designs, suggesting that the proposed self-supervised training paradigm is robust and compatible with diverse model architectures. Collectively, these results demonstrate that effective self-supervised pose-free learning can emerge from random initialization without relying on pretrained geometric priors.
}
\input{table/cross_dataset_generalization}

\subsection{Self-Supervised Pose-Free Learning}
\label{sec:results_with_priors}

\revision{
Pretrained geometric priors have become a common component in both supervised and self-supervised 3DGS methods, serving either to stabilize optimization~\cite{jiang2025anysplat,ye2025noposplat,hong2024pf3plat,kang2025selfsplat} or to improve reconstruction quality~\cite{chen2024mvsplat,charatan2024pixelsplat, xu2025depthsplat, ye2026yonosplat}. To facilitate a comprehensive and fair comparison with these approaches, we also investigate our framework under pretrained initialization. Importantly, these priors are used solely for initialization and are \textbf{not involved in the subsequent self-supervised optimization}.
For SPFSplatV2, the encoder, decoder, and Gaussian center head are initialized from pretrained MASt3R~\cite{leroy2024mast3r} weights, while the pose head is initialized to approximate the identity rotation matrix for stable convergence. For SPFSplatV2-L, the encoder, decoder, pose head, and Gaussian center head are initialized from pretrained VGGT~\cite{wang2025vggt} weights. All remaining layers are randomly initialized. Since the pretrained Gaussian center heads already provide a stable canonical parameterization, the additional canonical constraint introduced in Sec.~\ref{sec:ssl_from_scratch} is not required in this setting.
}

\textit{Novel View Synthesis:} 
Quantitative results on RE10K and ACID are reported in Tab.~\ref{tab:rek_results} and Tab.~\ref{tab:acid_results}. \revision{SPFSplatV2-L uses 224 $\times$ 224 inputs, while all other methods operate at 256 $\times$ 256}.  We make the following observations:
	1)	Despite being trained without ground-truth poses, SPFSplatV2 and SPFSplatV2-L consistently outperform state-of-the-art methods. Notably, both variants outperform our earlier SPFSplat baseline in most cases across different image overlap settings, with or without pose alignment, underscoring the effectiveness of the improved architecture.
	2)	While evaluation pose alignment generally improves performance, SPFSplatV2 without alignment still outperforms NoPoSplat with alignment, indicating that the jointly optimized poses are well aligned with the reconstructed Gaussians.
	3)	Between our variants, SPFSplatV2 achieves slightly higher PSNR on RE10K, while SPFSplatV2-L attains better LPIPS scores. On ACID, SPFSplatV2-L delivers the strongest overall results, likely benefiting from VGGT’s superior multi-view reconstruction capabilities and feature representations.
    4) Compared with NoPoSplat, which is initialized from the same MASt3R weights, SPFSplatV2 achieves significantly better performance. This result highlights the effectiveness of the proposed self-supervised optimization compared with NoPoSplat's reliance on ground-truth pose supervision during training.
    \revision{5) Compared with AnySplat, which relies on both VGGT initialization and VGGT-generated pseudo labels throughout training, SPFSplatV2-L achieves significantly better performance across all image-overlap settings while using the same initialization but without pseudo-label distillation. We attribute this improvement mainly to the difference in training supervision. AnySplat supervises rendering only on context images and therefore receives no direct supervision for novel viewpoints. In contrast, our framework supervises target-view renderings with self-predicted poses, resulting in improved NVS performance.}

Qualitative comparisons in Fig.~\ref{fig:qualitative_comparison} further demonstrate that our models reduce misalignment and recover more accurate geometry than baselines, even in challenging scenarios such as minimal input overlap or extreme viewpoint changes. Specifically, SPFSplatV2 improves structural accuracy and visual clarity compared to the original SPFSplat, while SPFSplatV2-L produces the highest overall rendering quality, capturing fine geometric details and textures more faithfully.

\input{figure/cross_dataset_generalization}

\textit{Cross-Dataset Generalization:}
To evaluate zero-shot generalization, we train on RE10K (indoor scenes) and test on ACID (outdoor), DTU (object-centric), DL3DV (outdoor), and ScanNet++ (indoor). As shown in Tab.~\ref{tab:cross-dataset}, both SPFSplatV2 variants generalize robustly across these diverse domains, consistently outperforming prior approaches. These datasets exhibit substantially different camera motions and scene types compared to RE10K, highlighting the strong out-of-domain generalization capability of our models, even under minimal image overlaps. Notably, in the RE10K$\rightarrow$ACID setting, SPFSplatV2 surpasses NoPoSplat and SPFSplat trained directly on ACID (Tab.~\ref{tab:acid_results}) when evaluated with pose alignment. SPFSplatV2-L consistently outperforms SPFSplatV2 both with and without pose alignment.

Qualitative results in Fig.~\ref{fig:cross_dataset} show that both variants produce sharper and more geometrically accurate reconstructions than prior methods, with SPFSplatV2-L achieving the highest visual quality. These results demonstrate that, even without ground-truth poses, our framework effectively aligns 3D Gaussians with predicted camera poses, enabling robust generalization to out-of-distribution scenes.

\input{table/pose_estimation_on_RE10k_and_ACID}

\input{table/scaling}

\textit{Relative Pose Estimation.}
We evaluate relative pose estimation between input image pairs on RE10K, ACID, DL3DV, and ScanNet++, with results in Tab.~\ref{tab:pose_estimation}. All splat-based methods are trained on RE10K to evaluate generalization. Since VGGT does not natively support 224 $\times$ 224 inputs, we  resize and center-crop its input images to 224 $\times$ 224 and pad the width to 518, as  specified in~\cite{wang2025vggt}. SPFSplatV2-L uses 224 $\times$ 224 inputs, while all other methods operate at 256 $\times$ 256. SuperPoint + SuperGlue derives relative poses from Essential Matrices estimated from feature correspondences. DUSt3R, MASt3R, and NoPoSplat use PnP\cite{hartley2003multiple} with RANSAC~\cite{fischler1981ransac}, while PF3plat and VGGT directly predict poses. Our SPFSplat variants support two strategies: (i) direct regression through the pose head, and (ii) PnP with RANSAC applied to predicted 3D Gaussian centers.

As shown in Tab.~\ref{tab:pose_estimation}, both regression and PnP yield similarly strong performance, reflecting consistent alignment between estimated poses and reconstructed 3D points. Despite no geometry priors during training, SPFSplatV2 substantially outperforms MASt3R, its initialization model, and SPFSplatV2-L improves over VGGT in most cases. This demonstrates our framework’s ability to jointly optimize camera poses and 3D structure using only image-level supervision. Both SPFSplatV2 variants also significantly surpass the original SPFSplat, primarily due to masked attention improving pose alignment.
On RE10K and ACID, our models achieve state-of-the-art results. On DL3DV and ScanNet++, which exhibit challenging camera motions, NoPoSplat achieves better performance, benefiting from ground-truth pose supervision during training. As shown in the scaling experiments below, this gap can be eliminated by scaling to larger training datasets.

\revision{
\textit{Scaling to Larger Training Datasets.}
One of the key advantages of the proposed training paradigm is its ability to leverage larger and more diverse image collections without requiring camera pose supervision during optimization. To evaluate this property, we progressively scale the training data from RE10K alone to a 2-dataset setting (RE10K+DL3DV) and a 4-dataset setting (RE10K+DL3DV+WildRGBD+CO3Dv2) in Tab.~\ref{tab:scaling}. As shown in Tab.~\ref{tab:scaling}, increasing the scale and diversity of the training data consistently improves both novel view synthesis and pose estimation performance across all evaluation datasets. For SPFSplatV2, training on two datasets leads to substantial gains on DL3DV as well as the out-of-domain WildRGBD and ScanNet++ benchmarks compared to training on RE10K alone. Similar improvements are observed for SPFSplatV2-L, with performance continuing to increase as the training data expands from two to four datasets. The consistent gains across both architectures suggest that the improvements stem from the proposed training paradigm rather than a particular network design. These results demonstrate that the proposed framework scales effectively with data and can continuously benefit from larger and more diverse training collections. This scalability is particularly attractive for exploiting internet-scale image and video data, where accurate camera poses are often unavailable.
}

\input{figure/3d_render_comparison}

\textit{Geometry Reconstruction:}
As illustrated in Fig.~\ref{fig:3d_vis}, SPFSplatV2 and SPFSplatV2-L produce substantially higher-quality 3D Gaussian primitives than prior methods, even under large viewpoint changes between input pairs. Previous approaches often exhibit distorted structures or ghosting artifacts, whereas our models, trained without ground-truth poses, reconstruct more accurate 3D geometry and yield sharper renderings, reflecting improved Gaussian alignment across views. This improvement arises from the joint optimization of Gaussians and poses, which strengthens geometric consistency. Compared to SPFSplat, SPFSplatV2 achieves more precise structural reconstruction, particularly visible in the left windows, while SPFSplatV2-L further enhances overall Gaussian quality.

\input{table/multi_view_extension}
\input{table/inference_efficiency}

\textit{Extension to Multiple Views:}
Our method naturally extends to multiple input views. As shown in Tab.~\ref{tab:multi_view_for_nvs}, novel view synthesis performance consistently improves with more context views. Both SPFSplatV2 and SPFSplatV2-L outperform the NoPoSplat and original SPFSplat, benefiting from enhanced architectures and the multi-view dropout strategy. With denser inputs, SPFSplatV2-L shows a clearer advantage, as its VGGT backbone, pretrained on multi-view data, provides stronger representations than MASt3R, which is limited to pairwise training. These results show that SPFSplatV2-L can better leverage additional views to enhance geometric consistency and reconstruction fidelity. Overall, the consistent gains with increasing context views highlight the flexibility and scalability of our framework for multi-view scenarios.

\subsection{Efficiency}
We compare the efficiency of our method to other approaches in Tab.~\ref{tab:inference_efficiency} and Tab.~\ref{tab:training_efficiency}.
\revision{All methods use 256$\times$ 256 inputs except SPFSplatV2-L, which uses 224$\times$224 due to its backbone architecture with a patch size of 14.}

\textit{Inference Efficiency:}
Tab.~\ref{tab:inference_efficiency} reports parameter size, FLOPs, and runtime during inference, measured for reconstructing 3D Gaussians from two input images on an A6000 GPU. 
SPFSplatV2 achieves comparable model size, FLOPs, and runtime to NoPoSplat and SPFSplat, while providing substantial speedups of approximately 3.5$\times$, 1.4$\times$, 2.3$\times$, and 27$\times$ over pixelSplat, MVSplat, SelfSplat, and PF3plat, respectively. These gains stem from architectural differences: pixelSplat relies on a time-consuming epipolar transformer; MVSplat requires costly cost-volume construction; and both SelfSplat and PF3plat depend on separate pose-estimation modules to lift predicted depth into Gaussians, with PF3plat further incurring heavy local feature matching costs. In contrast, SPFSplatV2 reconstructs Gaussians directly in a canonical space using a feed-forward network, avoiding explicit geometric operations such as cost-volume construction. Compared to SPFSplatV2, SPFSplatV2-L introduces additional computational overhead, as a trade-off for superior reconstruction quality.

\input{table/training_efficiency}

\textit{Training Efficiency:}
For a fair comparison, we evaluate the training efficiency of SPFSplat variants only against PF3plat and SelfSplat, as self-supervised pose-free methods require both context and target images during training, whereas other methods use only context images. The results are summarized in Tab.~\ref{tab:training_efficiency}. For all methods, two images are used as context views and one as the target view. Training time and GPU memory usage are measured on an NVIDIA A100 and averaged per sample. PF3plat requires significantly larger FLOPs, time, and GPU memory during training. Thanks to masked attention, SPFSplatV2 reduces training FLOPs of SPFSplat by 12\%, resulting in a 25\% speedup and a 13\% reduction in memory consumption. In contrast, SPFSplatV2-L incurs higher computational and memory costs, with a 46\% increase in FLOPs, 30\% longer training time, and 25\% higher memory usage compared to SPFSplat, reflecting the trade-off for its superior reconstruction performance. 

\subsection{Ablation Analysis}
\label{sec:experiments_ablation}

\input{table/component_ablation}
\input{figure/ablation}

\revision{
\textit{Ablation on Learnable Pose Token:}
As shown in Tab.~\ref{tab:ablation}, removing the learnable pose token and reverting to SPFSplat's global average pooling over feature maps leads to a marginal performance drop. This result suggests that the self-predicted pose estimation formulation is not tightly coupled to a specific pose prediction design. In particular, the learnable pose token enables compatibility with VGGT-style architectures while achieving slightly better performance.
}

\revision{
\textit{Ablation on Masked Attention:}
As shown in Tab.~\ref{tab:ablation}, replacing masked attention with SPFSplat’s two-branch input design results in consistent performance degradation across both pose estimation and novel view synthesis, with a particularly significant impact on pose estimation accuracy. This effect is further illustrated in Fig.~\ref{fig:ablation}, where SPFSplat exhibits significant pose drift, whereas SPFSplatV2 produces camera poses that are better aligned with the reconstructed 3D geometry, leading to more accurate renderings.
}

\revision{
\textit{Ablation on Intrinsic Embeddings:}
As shown in Tab.~\ref{tab:ablation}, removing intrinsic embeddings from the backbone reduces both reconstruction and pose estimation accuracy. This degradation is primarily attributed to increased scale ambiguity during joint optimization of 3D Gaussians and camera poses. Nevertheless, even without intrinsic embeddings, our method still outperforms NoPoSplat with intrinsic embeddings (Tab.~\ref{tab:rek_results}), achieving improvements of 0.5dB in PSNR when both evaluated with pose alignment.
}

\revision{
\textit{Ablation on Reprojection Loss:}
Finally, removing the reprojection loss in Tab.~\ref{tab:ablation} and optimizing solely with the target-view rendering loss leads to a substantial degradation in both novel view synthesis and pose estimation performance. This highlights the importance of enforcing geometric consistency between 3D points and camera poses, and demonstrates that reprojection supervision is a key component for stable joint optimization and accurate reconstruction.
}

\input{figure/ablation_multiview_dropout}

\input{table/gt_poses_ablation}

\revision{
\textit{Ablation on Multi-view Dropout:}
As in Fig.~\ref{fig:multiview_dropout}, under the 5-view training setting, multi-view dropout randomly removes context views during training, exposing the model to diverse view configurations with up to five input views. In contrast, the baseline without multi-view dropout is trained using a fixed 5-view configuration. As a result, multi-view dropout enables the model to effectively generalize to varying numbers of input views at test time, with performance consistently improving as additional views are provided. Notably, the model continues to benefit from up to 10 input views during evaluation, despite never being trained with more than five views. These results demonstrate that multi-view dropout improves robustness to varying view counts and spatial distributions, leading to better generalization across multi-view settings.
}

\textit{Ablation on Ground-Truth Poses:}
In Tab.~\ref{tab:gt_poses_ablation}, we evaluate our framework’s ability to reconstruct geometry without ground-truth pose supervision by analyzing the effect of incorporating ground-truth poses during training in two settings: (b) rendering novel views using ground-truth poses, as in NoPoSplat, and (c) introducing a pose loss that penalizes the discrepancy between predicted and ground-truth poses, while still rendering with predicted poses. The pose loss combines a geodesic loss\cite{salehi2018geo} for rotation and an $L_2$ loss for translation. Since ground-truth poses are used only during training, the framework remains pose-free at inference.

Relative to setting (b), (a) SPFSplatV2 achieves improvements in both novel view synthesis and pose estimation. This can be attributed to the joint optimization of Gaussians and poses, which encourages better geometric alignment and more consistent feature learning. From (a) to (c), adding a pose loss improves pose accuracy but yields only marginal gains in novel view synthesis, underscoring the model’s capacity to reconstruct geometry without explicit pose supervision. These findings also suggest that high-quality novel view synthesis depends on factors beyond pose accuracy, such as occlusion, textureless regions, and extreme viewpoint changes, which may require generative priors or explicit 3D supervision.

\textit{Evaluation on In-the-Wild Data:}
We highlight the effectiveness of our model on mobile phone photos using SPFSplatV2 without intrinsic embeddings. 
 The 3D geometry and rendered results in Fig.~\ref{fig:phone_vis} demonstrate strong out-of-domain generalization, even under large viewpoint changes.

\textit{Failure Cases:}
As shown in Fig.~\ref{fig:bad_cases}, our method can produce blurred outputs or artifacts in occluded or texture-less regions, or under extreme viewpoint changes. Addressing these limitations may require stronger generative capabilities or larger training data.
\input{figure/phone_image}

%% file: table/nvs_from_scratch.tex
\begin{table*}[ht]
      \caption{\revision{Performance comparison under random initialization. Methods in the upper block use pretrained initialization and are included for reference. All methods except AnySplat are trained on RE10K. AnySplat is trained on large-scale mixed datasets. The \textbf{best} results are highlighted and $\ast$ denotes evaluation with pose alignment.}}
\label{tab:nvs-from-scratch}
    \setlength{\tabcolsep}{11pt}
    \centering

    \begin{tabular}{l ccccc ccccc}
        \toprule
        \multirow{2}{*}{\textbf{Method}} & \multicolumn{5}{c}{\textbf{RE10K}} & \multicolumn{5}{c}{\textbf{ACID}} \\
        \cmidrule(lr){2-6} \cmidrule(lr){7-11}          
        &  PSNR$\uparrow$ & SSIM$\uparrow$ & LPIPS$\downarrow$ & 
        10$^\circ$ $\uparrow$ & 20$^\circ$ $\uparrow$ &
        PSNR$\uparrow$ & SSIM$\uparrow$ & LPIPS$\downarrow$  & 
        10$^\circ$ $\uparrow$ & 20$^\circ$ $\uparrow$ \\
        \midrule
        \rowcolor{gray!20} \multicolumn{11}{l}{\textit{Pretrained Init.}} \\ 
    \graycell{SelfSplat} &
    \graycell{19.152} & \graycell{0.680} & \graycell{0.328}
    & \graycell{0.184} & \graycell{0.318}
    & \graycell{22.204} & \graycell{0.686} & \graycell{0.316}
    & \graycell{0.258} & \graycell{0.396}\\
    
    \graycell{PF3plat} &
\graycell{21.042} & \graycell{0.739} & \graycell{0.233}
& \graycell{0.398} & \graycell{0.613}
& \graycell{20.726} & \graycell{0.610} & \graycell{0.308}
& \graycell{0.165} & \graycell{0.340}\\

    \graycell{AnySplat} &
    \graycell{18.039} & \graycell{0.643} & \graycell{0.324}
    & \graycell{0.584} & \graycell{0.740}
    & \graycell{19.908} & \graycell{0.596} & \graycell{0.346}
    & \graycell{0.436} & \graycell{0.607}\\
        \rowcolor{gray!20} \multicolumn{11}{l}{\textit{Random Init.}} \\ 
        NoPoSplat$^\ast$ & 21.382 & 0.709 & 0.266
        & 0.014 & 0.055 
        & 23.193 & 0.667 & 0.274
        & 0.017 & 0.072 \\
        SPFSplat & 21.306 & 0.693 & 0.248 
        & 0.239 & 0.398 
        & 23.354 & 0.662 & 0.260 
        & 0.182 & 0.344 \\
        SPFSplatV2 & 22.383 & 0.725 & \textbf{0.223}
        & 0.422 & 0.572 
        & 24.267& \textbf{0.705} & 0.243
        & 0.289	& 0.429 \\
        SPFSplatV2-L & \textbf{22.435} & \textbf{0.737} & {0.231} 
        & \textbf{0.439} & \textbf{0.583} 
        & \textbf{24.299} & {0.702} & \textbf{0.229} 
        & \textbf{0.305} &	\textbf{0.443} \\
        \bottomrule
    \end{tabular}

\end{table*}

%% file: table/novel_view_synthesis_comparison_on_RE10k.tex
\begin{table*}[!ht]
    \caption{\revision{Performance comparison of two-view novel view synthesis on RE10K~\cite{zhou2018stereo} with different image overlap. Our method outperforms state-of-the-art approaches.  
    The \textbf{best} and \underline{second-best} results are highlighted, and $\ast$ denotes evaluation with pose alignment.}}
    \label{tab:rek_results}
    \centering
    \resizebox{\textwidth}{!}{ 
    \begin{tabular}{lcccccccccccc}
    \toprule
     \multirow{2}{*}{\textbf{Method}}  & \multicolumn{3}{c}{\textbf{Small}} & \multicolumn{3}{c}{\textbf{Medium}} & \multicolumn{3}{c}{\textbf{Large}} & \multicolumn{3}{c}{\textbf{Average}} \\
    \cmidrule(lr){2-4} \cmidrule(lr){5-7}              \cmidrule(lr){8-10} \cmidrule(lr){11-13} 
    & PSNR $\uparrow$ & SSIM $\uparrow$ & LPIPS $\downarrow$ 
    & PSNR $\uparrow$ & SSIM $\uparrow$ & LPIPS $\downarrow$ 
    & PSNR $\uparrow$ & SSIM $\uparrow$ & LPIPS $\downarrow$ 
    & PSNR $\uparrow$ & SSIM $\uparrow$ & LPIPS $\downarrow$ \\ 
    \midrule
    \rowcolor{gray!20} \multicolumn{13}{l}{\textit{Pose-Required}} \\ 
    pixelSplat & 20.277 & 0.719 & 0.265 
    & 23.726 & 0.811 & 0.180 
    & 27.152 & 0.880 & 0.121 
    & 23.859 & 0.808 & 0.184  \\
    MVSplat & 20.371 & 0.725 & 0.250  
    & 23.808 & 0.814 & 0.172 
    & 27.466 & 0.885 & 0.115
    & 24.012 & 0.812 & 0.175  \\
    \revision{DepthSplat} & 22.820 & 0.798 & 0.193 
    & 25.383 & 0.851 & 0.145
    & 28.317 & 0.900 & 0.104 
    & 25.595 & 0.852 & 0.145 \\
    \revision{YoNoSplat} & 22.199 & 0.768 & 0.205 
    & 24.156 & 0.813 & 0.161 
    & 26.103 & 0.850 & 0.128 
    & 24.233 & 0.813 & 0.162 \\
    \midrule
    \rowcolor{gray!20} \multicolumn{13}{l}{\textit{Supervised Pose-Free}} \\ 
    CoPoNeRF & 17.393 & 0.585 & 0.462 
    & 18.813 & 0.616 & 0.392 
    & 20.464 & 0.652 & 0.318 
    & 18.938 & 0.619 & 0.388 \\
    Splatt3R & 17.789 & 0.582 & 0.375 
    & 18.828 & 0.607 & 0.330 
    & 19.243 & 0.593 & 0.317 
    & 18.688 & 0.337 & 0.596  \\
    NoPoSplat$^\ast$ & 22.514 & 0.784 & 0.210 
    & 24.899 & 0.839 & 0.160 
    & 27.411 & 0.883 & 0.119 
    & 25.033 & 0.838 & 0.160 \\
    \midrule
    \rowcolor{gray!20} \multicolumn{13}{l}{\textit{Self-Supervised Pose-Free}} \\ 
    SelfSplat & 14.828 & 0.543 & 0.469 
    & 18.857 & 0.679 & 0.328 
    & 23.338 & 0.798 & 0.208
    & 19.152 & 0.680 & 0.328 \\
    PF3plat & 18.358 & 0.668 & 0.298
    & 20.953 & 0.741 & 0.231 
    & 23.491 & 0.795 & 0.179 
    & 21.042 & 0.739 & 0.233  \\
     \revision{AnySplat} & 15.540 & 0.601 & 0.393 
    & 17.869 & 0.640 & {0.325}
    &  {20.459} &  {0.685} &  {0.260} 
    &  {18.039} &  {0.643} &  {0.324} \\
    
    SPFSplat & 22.897 &	0.792 &	0.201
    & 25.334 & 0.847 & 0.153
    & 27.947 & 0.894 & 0.110
    & 25.484 & 0.847 & 0.153 \\
    SPFSplat$^\ast$  & 23.178 & 0.796 & 0.200
    & 25.695 & 0.853 & 0.151
    & 28.377 & 0.899 & 0.111
    & 25.845 & 0.852 & 0.152 \\

    \textbf{SPFSplatV2} & 23.123 & 0.800 & 0.195
    & 25.542 & 0.853 & 0.149
    & 28.143 & 0.897 & 0.110
    & 25.693 & 0.853 & 0.149 \\  
    \textbf{SPFSplatV2$^\ast$} & \textbf{23.456} & \textbf{0.806} & 0.193	
    & \textbf{26.030} & \textbf{0.862}	& 0.145
    & \textbf{28.682} & \textbf{0.905} & 0.107
    & \textbf{26.157} & \textbf{0.861} & 0.146 \\ 
    \textbf{SPFSplatV2-L} & 23.138 & \underline{0.804} & \underline{0.184}
    & 25.518 & 0.856 & \underline{0.136}
    & 28.081 & 0.899 & \underline{0.099}
    & 25.668 & 0.855 & \underline{0.137} \\
    \textbf{SPFSplatV2-L$^\ast$} & \underline{23.329} & \underline{0.804} & \textbf{0.183}	
    & \underline{25.863} & \underline{0.861} & \textbf{0.134}
    & \underline{28.456} & \underline{0.903} & \textbf{0.098}
    & \underline{25.983} & \underline{0.859} & \textbf{0.136} \\
    \bottomrule
    \end{tabular}
    }
    
\end{table*}

%% file: table/novel_view_synthesis_comparison_on_ACID.tex
\begin{table*}[!ht]
\footnotesize
\caption{Performance comparison of two-view novel view synthesis on ACID~\cite{liu2021infinite}. The \textbf{best} and \underline{second best} results are highlighted.}
\centering
\resizebox{\textwidth}{!}{ 

\begin{tabular}{lcccccccccccc}
\toprule
 \multirow{2}{*}{\textbf{Method}}  & \multicolumn{3}{c}{\textbf{Small}} & \multicolumn{3}{c}{\textbf{Medium}} & \multicolumn{3}{c}{\textbf{Large}} & \multicolumn{3}{c}{\textbf{Average}} \\
    \cmidrule(lr){2-4} \cmidrule(lr){5-7}              \cmidrule(lr){8-10} \cmidrule(lr){11-13} 
& PSNR $\uparrow$ & SSIM $\uparrow$ & LPIPS $\downarrow$ 
& PSNR $\uparrow$ & SSIM $\uparrow$ & LPIPS $\downarrow$ 
& PSNR $\uparrow$ & SSIM $\uparrow$ & LPIPS $\downarrow$ 
& PSNR $\uparrow$ & SSIM $\uparrow$ & LPIPS $\downarrow$ \\ 
\midrule
\rowcolor{gray!20} \multicolumn{13}{l}{\textit{Pose-Required}} \\ 
pixelSplat & 22.088 & 0.655 & 0.284
& 25.525 & 0.777 & 0.197
& 28.527 & 0.854 & 0.139 
& 25.889 & 0.780 & 0.194 \\
MVSplat & 21.412 & 0.640 & 0.290  
& 25.150 & 0.772 & 0.198
& 28.457 & 0.854 & 0.137
& 25.561 & 0.775 & 0.195 \\
\midrule
\rowcolor{gray!20} \multicolumn{13}{l}{\textit{Supervised Pose-Free}} \\ 
CoPoNeRF & 18.651 & 0.551 & 0.485
& 20.654 & 0.595 & 0.418 
& 22.654 & 0.652 & 0.343 
& 20.950 & 0.606 & 0.406 \\
Splatt3R & 17.419 & 0.501 & 0.434 
& 18.257 & 0.514 & 0.405 
& 18.134 & 0.508 & 0.395 
& 18.060 & 0.510 & 0.407 \\
NoPoSplat$^\ast$ & 23.087 & 0.685 & 0.258 
& 25.624 & 0.777 & 0.193 
& 28.043 & 0.841 & 0.144 
& 25.961 & 0.781 & 0.189 \\
\midrule
\rowcolor{gray!20} \multicolumn{13}{l}{\textit{Self-Supervised Pose-Free}} \\ 
SelfSplat & 18.301 & 0.568 & 0.408
& 21.375 & 0.676 & 0.314 
& 25.219 & 0.792 & 0.214
& 22.089 & 0.694 & 0.298 \\
PF3plat & 18.112 & 0.537 & 0.376 
& 20.732 & 0.615 & 0.307 
& 23.607 & 0.710 & 0.228
& 21.206 & 0.632 & 0.293 \\

SPFSplat & 22.667 &	0.665 & 0.262
& 25.620	& 0.773	& 0.192
& 28.607 & 0.856 & 0.136
& 26.070	& 0.781	& 0.186 \\
SPFSplat$^\ast$  & \underline{23.676} & \underline{0.708} & 0.243
& 26.351 & \underline{0.801} & 0.182
& 29.170	& \underline{0.870} & 0.131
& 26.796 & \underline{0.807} & 0.176 \\

\textbf{SPFSplatV2} & 22.944 & 0.679 & 0.255
& 25.849 & 0.784 & 0.187
& 28.766 & 0.862 & 0.133
& 26.284 & 0.791 & 0.182 \\
\textbf{SPFSplatV2$^\ast$} & 23.635	& 0.700 & 0.247	
& \underline{26.356} & 0.798 & 0.182
& \textbf{29.223} & \textbf{0.871} & 0.129
& \underline{26.809} & 0.804 & 0.176 \\
\textbf{SPFSplatV2-L} & {23.640}	& {0.706}	& \underline{0.225}
& 26.272 & \underline{0.801} & \underline{0.166}
& 28.938 & {0.868} & \underline{0.120}
& 26.674 & {0.806} & \underline{0.162} \\
\textbf{SPFSplatV2-L$^\ast$} & \textbf{23.937} & \textbf{0.710} & \textbf{0.224}	
& \textbf{26.489} & \textbf{0.803} & \textbf{0.165}
& \underline{29.188} & \textbf{0.871} & \textbf{0.118} 
& \textbf{26.917} & \textbf{0.809} & \textbf{0.160}\\
\bottomrule
\end{tabular}
}

\label{tab:acid_results}
\end{table*}

%% file: figure/qualitative_comparison_on_RE10k_and_ACID.tex
\begin{figure*}[ht]
    \centering
   
    \includegraphics[width=1.0\textwidth]{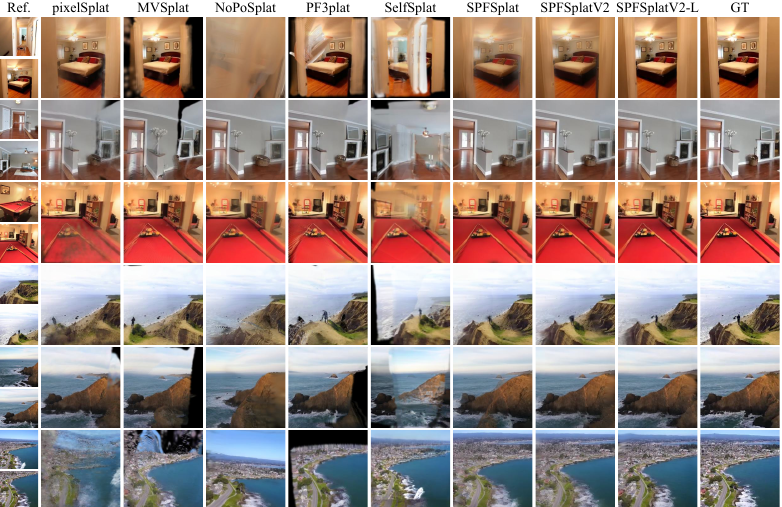}

    \caption{Qualitative comparison on RE10K (top three rows) and ACID (bottom three rows). 
    Our method 1) better handles extreme viewpoint changes and minimal input overlap (e.g., Row~1 and Row~2), 
    2) preserves finer details (e.g., Row~3) and more accurate geometric structure (e.g., Row~4 and Row~5), and 
    3) reduces misaligned blending artifacts and ghosting effect (e.g. Row~5 and Row~6).
    }

    \label{fig:qualitative_comparison}
\end{figure*}


%% file: table/cross_dataset_generalization.tex
\begin{table*}[ht]
        \caption{\revision{Performance comparison of cross-dataset generalization on ACID, DTU, DL3DV, and ScanNet++. All methods are trained on RE10K, except  \graycell{AnySplat}, which is trained on mixed datasets including DL3DV and ScanNet++; we nevertheless report its in-domain performance on these datasets for completeness. Our method demonstrates superior cross-dataset generalization over state-of-the-art approaches. $\ast$ denotes evaluation with pose alignment.}}
\label{tab:cross-dataset}
    \setlength{\tabcolsep}{6.5pt}
    \centering

    \begin{tabular}{l ccc ccc ccc ccc}
        \toprule
        \multirow{2}{*}{\textbf{Method}} & \multicolumn{3}{c}{\textbf{ACID}} & \multicolumn{3}{c}{\textbf{DTU}} &
        \multicolumn{3}{c}{\textbf{DL3DV}} &
        \multicolumn{3}{c}{\textbf{ScanNet++}} \\
        \cmidrule(lr){2-4} \cmidrule(lr){5-7}              \cmidrule(lr){8-10} \cmidrule(lr){11-13} 
        &  PSNR$\uparrow$ & SSIM$\uparrow$ & LPIPS$\downarrow$ & PSNR$\uparrow$ & SSIM$\uparrow$ & LPIPS$\downarrow$  &
        PSNR$\uparrow$ & SSIM$\uparrow$ & LPIPS$\downarrow$ &
        PSNR$\uparrow$ & SSIM$\uparrow$ & LPIPS$\downarrow$ \\
        \midrule
        \rowcolor{gray!20} \multicolumn{13}{l}{\textit{Pose-Required}} \\
        pixelSplat & 25.477 & 0.770 & 0.207 &
        15.067 & 0.539 & 0.341 &
        18.688 & 0.582 & 0.354 &
        18.422 & 0.720 & 0.278  \\
         MVSplat & 25.525 & 0.773 & 0.199 &
         14.542 & 0.537 & 0.324 &
         17.786 & 0.545 & 0.357 &  
         17.138 & 0.687 & 0.297\\
         \revision{DepthSplat} & 26.012 & 0.791 & 0.185 
         & 12.384 & 0.404 & 0.498 
         & 19.553 & 0.611 & 0.285 
         & 20.775 & 0.760 & 0.254 \\
         \revision{YoNoSplat} & 24.246 & 0.721 & 0.222 
         & 12.532 & 0.342 & 0.532
         & 19.636 & 0.594 & 0.311
         & 21.075 & 0.744 & 0.254 \\
         \midrule
         \rowcolor{gray!20} \multicolumn{13}{l}{\textit{Supervised Pose-Free}} \\
         NoPoSplat$^\ast$ & 25.764 & 0.776 & 0.199 &
         {17.899} & {0.629} & {0.279} &
         19.974 & 0.612 & 0.305 &
         22.136 & 0.798 & 0.232\\
         \midrule
         \rowcolor{gray!20} \multicolumn{13}{l}{\textit{Self-Supervised Pose-Free}} \\
         SelfSplat & 22.204 & 0.686 & 0.316 &
         13.249 & 0.434 & 0.441 &
         15.047 & 0.410 & 0.498 &
          13.277 & 0.538 & 0.534 \\
         PF3plat & 20.726 & 0.610 & 0.308 &
         12.972 & 0.407 & 0.464 &
         15.773 & 0.458 & 0.417 &
         16.471 & 0.688 & 0.303 \\
         \revision{AnySplat} & 19.908 & 0.596 & 0.346 &
         11.990 & 0.410 & 0.450 & 
         \graycell{14.621} &  \graycell{0.442} &  \graycell{0.428} & 
          \graycell{15.804} &  \graycell{0.661} &  \graycell{0.358}\\
         SPFSplat & 25.965 & 0.781 & 0.190
         & 16.550 & 0.579	& 0.270
         & 19.172 & 0.573 & 0.315
         & 19.971 & 0.738 & 0.265 \\
         SPFSplat$^\ast$& \underline{26.697}	& \textbf{0.806}	& 0.181
         & 18.297 & 0.660	& 0.255
         & 19.494 & 0.574 & 0.319
         & 22.312 & 0.793 & 0.243 \\
          \textbf{SPFSplatV2} & 26.220	& 0.789	& 0.185	
          & 16.793 & 0.584 &0.265
          & 19.439 & 0.584 & 0.304
          & 20.919 & 0.771 & 0.243 \\
         \textbf{SPFSplatV2$^\ast$} & \textbf{26.802}	& \underline{0.805} &0.179
         & \underline{18.506} & \underline{0.663} & 0.246
         & \underline{19.978} & 0.607 & 0.302
         & \underline{22.776} & \underline{0.812} & 0.227\\
         \textbf{SPFSplatV2-L} & 26.361 & 0.796 & \underline{0.169}	& 17.739 & 0.653 & \textbf{0.228}
         & 19.743 & \underline{0.613} & \textbf{0.277}
         & 21.796 & 0.811 & \underline{0.200} \\
        \textbf{SPFSplatV2-L$^\ast$} & 26.680 & 0.802 &	\textbf{0.166}	
        & \textbf{19.316} & \textbf{0.671} & \underline{0.229}
        & \textbf{20.108} & \textbf{0.615} & \underline{0.279}
        & \textbf{23.072} & \textbf{0.820} & \textbf{0.199} \\
        \bottomrule
    \end{tabular}

\end{table*}

%% file: figure/cross_dataset_generalization.tex
\begin{figure*}[ht]
    \centering
   
    \includegraphics[width=1.0\textwidth]{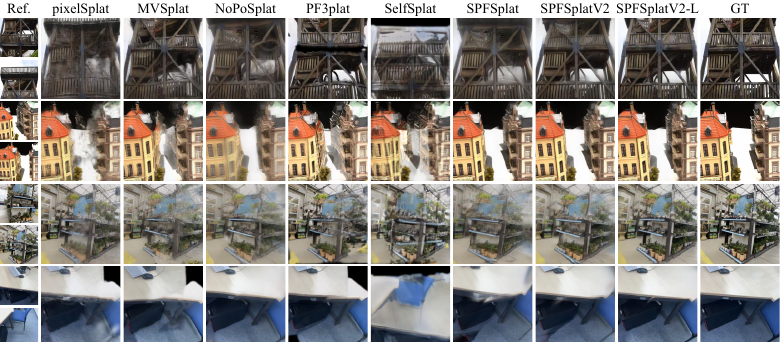}
    
    \caption{Qualitative comparison on cross-dataset generalization. All methods are trained on
    RE10K and evaluated on ACID and DTU, DL3DV, and ScanNet++ (from top to bottom). 
    Both SPFSplatV2 and SPFSplatV2-L yield more geometrically accurate reconstructions than prior methods.
    }
    \label{fig:cross_dataset}
\end{figure*}

%% file: table/pose_estimation_on_RE10k_and_ACID.tex
\begin{table*}[!ht]
\caption{\revision{Performance comparison of pose estimation in AUC with various thresholds on RE10K, ACID, DL3DV and ScanNet++ datasets. All NVS-based methods are trained on RE10K, except \textcolor{lightgray}{AnySplat}, which is trained on mixed datasets including DL3DV and ScanNet++; we nevertheless report its in-domain performance on these datasets for completeness.}}
\label{tab:pose_estimation}
\centering
\setlength{\tabcolsep}{8pt}

\begin{tabular}{l ccc ccc ccc ccc}
        \toprule
        \multirow{2}{*}{\textbf{Method}} & \multicolumn{3}{c}{\textbf{RE10K}} & \multicolumn{3}{c}{\textbf{ACID}} &
        \multicolumn{3}{c}{\textbf{DL3DV}} &
        \multicolumn{3}{c}{\textbf{ScanNet++}}   \\
    \cmidrule(lr){2-4} \cmidrule(lr){5-7} \cmidrule(lr){8-10} \cmidrule(lr){11-13}  
    &  5$^\circ$ $\uparrow$ & 10$^\circ$ $\uparrow$ & 20$^\circ$ $\uparrow$ 
    &  5$^\circ$ $\uparrow$ &  10$^\circ$ $\uparrow$ &  20$^\circ$ $\uparrow$ 
    &  5$^\circ$ $\uparrow$ & 10$^\circ$ $\uparrow$ & 20$^\circ$ $\uparrow$ 
    &  5$^\circ$ $\uparrow$ & 10$^\circ$ $\uparrow$ & 20$^\circ$ $\uparrow$   \\
    \midrule
    \rowcolor{gray!20} \multicolumn{13}{l}{\textit{SfM-based Methods}} \\
    SP + SG & 0.234 & 0.406 & 0.569 
    & 0.228 & 0.363 & 0.500 
    & 0.224 & 0.372 & 0.492 
    & 0.087 & 0.151 & 0.248 \\
    DUSt3R & 0.336 & 0.541 & 0.702 
    & 0.118 & 0.279 & 0.470 
    & 0.275 & 0.490 & 0.686 
    & 0.109 & 0.284 & 0.500 \\ 
    MASt3R & 0.281 & 0.494 & 0.672 
    & 0.138 & 0.312 & 0.507 
    & 0.332 & 0.593 & 0.772 
    & 0.139 & 0.336 & 0.549  \\
    VGGT & 0.257 & 0.474 & 0.658 
    & 0.142	& 0.304	& 0.486
    & 0.356	& \underline{0.609}	& \textbf{0.784}	
    & 0.156	& 0.311	& 0.514\\
    \rowcolor{gray!20} \multicolumn{13}{l}{\textit{NVS-based Methods}} \\
    NoPoSplat & 0.571 & 0.727 & 0.833
    & 0.335 & 0.496 & 0.644
    & \textbf{0.470} & \textbf{0.646} & \underline{0.762} 
    & \textbf{0.207} & \textbf{0.403} & \textbf{0.641} \\
    PF3plat  & 0.187 & 0.398 & 0.613 
    & 0.060 & 0.165 & 0.340 
    & 0.118 & 0.281 & 0.479 
    & 0.058 & 0.204 & 0.415  \\
    \revision{AnySplat} & 0.375 & 0.584 & 0.740 & 
    0.249 & 0.436 & 0.607 & 
    \textcolor{lightgray}{0.368} & \textcolor{lightgray}{0.625} & \textcolor{lightgray}{0.800} &
    \textcolor{lightgray}{0.182} & \textcolor{lightgray}{0.428} & \textcolor{lightgray}{0.670} \\
    {SPFSplat} & 0.617	& 0.755	& 0.845
    & 0.364	& 0.520 & 0.662
    & 0.283	& 0.461	& 0.622
    & 0.098	& 0.188	& 0.374 \\
    {SPFSplat (PnP)} & 0.613 & 0.754 & 0.845
    & 0.355	& 0.516	& 0.658
    & 0.279	& 0.464	& 0.626
    & 0.120 & 0.226 & 0.408 \\
     \textbf{SPFSplatV2 } & 0.638 & 0.776 & 0.863	
     & \textbf{0.387}	& \textbf{0.541}	& \textbf{0.672}
     & 0.369	& 0.534	& 0.694
     &  0.111	& 0.250 & 0.463 \\
     \textbf{SPFSplatV2 (PnP)} & 0.641 & 0.777 & \underline{0.864}	& 0.374	& 0.533	& 0.667
     & 0.375	& 0.542	& 0.700
     & 0.144	& 0.281	& 0.487  \\
     \textbf{SPFSplatV2-L } & \underline{0.645} & \underline{0.780} & \underline{0.864}	
     & \underline{0.379}	& \underline{0.539}	& \underline{0.671}
     & 0.420 & 0.582 & 0.711
     & \underline{0.184}	& \underline{0.400} & \underline{0.630} \\
     \textbf{SPFSplatV2-L (PnP)} & \textbf{0.657}	& \textbf{0.786}	& \textbf{0.867}	& {0.375}	& {0.535}	& {0.668}
     & \underline{0.429}	& {0.587}	& {0.716}
     & 0.183	& \underline{0.400} & 0.627 \\

    \bottomrule
    \end{tabular}

\end{table*}

%% file: table/scaling.tex
\begin{table*}[!ht]
\caption{\revision{Scaling SPFSplatV2 and SPFSplatV2-L to larger and more diverse training datasets. The 2-dataset setting corresponds to RE10K+DL3DV, while the 4-dataset setting corresponds to RE10K+DL3DV+WildRGBD+CO3Dv2. Increasing the scale and diversity of the training data consistently improves both novel view synthesis and pose estimation performance. }}
\label{tab:scaling}
\centering
\setlength{\tabcolsep}{6.3pt}

\begin{tabular}{l c cccc cccc cccc}
        \toprule
        \multirow{2}{*}{\textbf{Method}} & 
        \multirow{2}{*}{\textbf{Training Data}} & 
        \multicolumn{4}{c}{\textbf{WildRGBD}} & \multicolumn{4}{c}{\textbf{ScanNet++}} &
        \multicolumn{4}{c}{\textbf{DL3DV}}   \\
    \cmidrule(lr){3-6} \cmidrule(lr){7-10} \cmidrule(lr){11-14}  
    & &  PSNR & 5$^\circ$ $\uparrow$ & 10$^\circ$ $\uparrow$ & 20$^\circ$ $\uparrow$ 
    &  PSNR & 5$^\circ$ $\uparrow$ &  10$^\circ$ $\uparrow$ &  20$^\circ$ $\uparrow$ 
    &  PSNR &5$^\circ$ $\uparrow$ & 10$^\circ$ $\uparrow$ & 20$^\circ$ $\uparrow$  \\
    \midrule
    \multirow{2}{*}{\textbf{SPFSplatV2}} & RE10K & 22.609 & 0.392 & 0.654 & 0.808
    & 20.919 & 0.111 & 0.250 & 0.463
    & 19.439 & 0.369 & 0.534 & 0.694\\
    & 2 datasets 
    & \textbf{23.482} & \textbf{0.487} & 
    \textbf{0.713} & \textbf{0.842}
    & \textbf{22.934} & \textbf{0.251} & \textbf{0.478} & \textbf{0.698} 
    & \textbf{20.938} & \textbf{0.560} & \textbf{0.711} &  \textbf{0.806}\\
    \midrule
    \multirow{3}{*}{\textbf{SPFSplatV2-L}} & RE10K 
    & 22.939 & 0.386 & 0.650	& 0.807
    & 21.796 & 0.184 & 0.400 & 0.630	
    & 19.743 & 0.420 &	0.582 & 0.711\\
    & 2 datasets 
    & 23.800	& 0.534 &	0.740 &	0.855	
    & 23.522 & 0.268 & 0.473 & 0.670	
    & 20.886 & 0.568 & 0.713 & 0.809 \\
    & 4 datasets 
    & \textbf{24.079} & \textbf{0.610} &	\textbf{0.774} &	\textbf{0.873}	
    & \textbf{23.554} &	\textbf{0.279} &	\textbf{0.477} &	\textbf{0.676}
    & \textbf{20.984} & \textbf{0.576} & \textbf{0.719} & \textbf{0.815} \\
    \bottomrule
    \end{tabular}

\end{table*}

%% file: figure/3d_render_comparison.tex



\begin{figure*}[!ht]
    \centering
    \includegraphics[width=\linewidth]{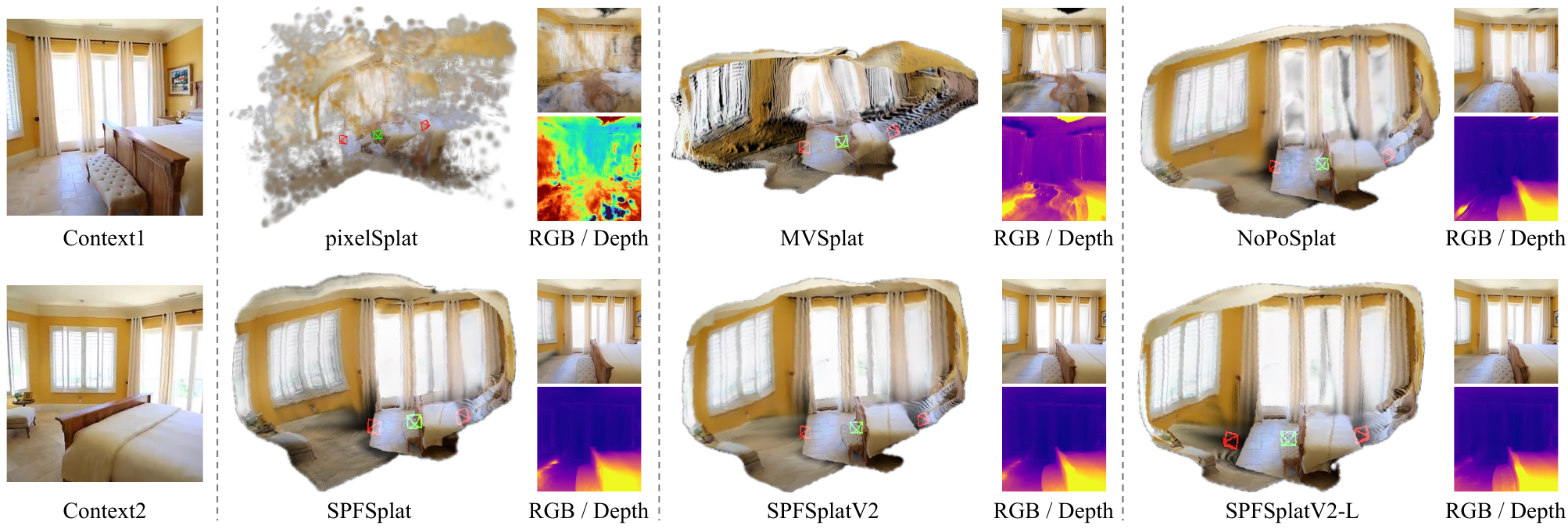}
    \caption{Comparison of 3D Gaussians and rendered results. Red and green denote context and target camera poses, respectively. Rendered images and depth maps at the target views are shown on the right. Our method produces higher-quality 3D Gaussians and better rendering over baselines.}
    \label{fig:3d_vis}
\end{figure*}

%% file: table/multi_view_extension.tex
\begin{table*}[ht]
\footnotesize
\caption{Novel view synthesis with varying input view numbers. For NoPoSplat, only results reported in~\cite{ye2025noposplat} are shown.}
\label{tab:multi_view_for_nvs}
\setlength{\tabcolsep}{7pt}
    \centering
    \begin{tabular}{l ccc ccc ccc ccc}
        \toprule
        \multirow{2}{*}{\textbf{Method}} & 
        \multicolumn{3}{c}{2 Views} & 
        \multicolumn{3}{c}{3 Views} & \multicolumn{3}{c}{5 Views} &
        \multicolumn{3}{c}{10 Views}\\
        \cmidrule(lr){2-4} \cmidrule(lr){5-7}
        \cmidrule(lr){8-10} \cmidrule(lr){11-13} 
        & PSNR$\uparrow$ & SSIM$\uparrow$ & LPIPS$\downarrow$ & PSNR$\uparrow$ & SSIM$\uparrow$ & LPIPS$\downarrow$  &PSNR$\uparrow$ & SSIM$\uparrow$ & LPIPS$\downarrow$ &PSNR$\uparrow$ & SSIM$\uparrow$ & LPIPS$\downarrow$   \\
        \midrule
        NoPoSplat$^\ast$ & 25.033 & 0.838 & 0.160 & 26.619 & 0.872 & 0.127 & -- & -- & -- & -- & -- & -- \\
        SPFSplat & 25.484 & 0.847 & 0.153 & 26.724 &  0.871  & 0.128 & 26.891 & 0.875 & 0.122 & 27.159  & 0.880 & 0.115 \\
        SPFSplatV2 & \textbf{25.693}	& \underline{0.853}	& \underline{0.149} & \underline{27.262}	& \underline{0.884}	& \underline{0.120} & \underline{27.585} & \underline{0.890} & \underline{0.115} & \underline{28.188}	& \underline{0.901} & \underline{0.106}  \\
        SPFSplatV2-L & \underline{25.668} & \textbf{0.855} & \textbf{0.137} & \textbf{27.685} & \textbf{0.898} & \textbf{0.101} & \textbf{28.141} & \textbf{0.907} & \textbf{0.094} & \textbf{28.973} &	\textbf{0.922} & \textbf{0.083} \\
        \bottomrule
    \end{tabular}
\end{table*}

%% file: table/inference_efficiency.tex


\begin{table}[ht]
\caption{Comparison of inference Efficiency on An NVIDIA A6000 GPU. }
\setlength{\tabcolsep}{4.5pt}
\label{tab:inference_efficiency}
    \centering
    \begin{tabular}{lccc}
        \toprule
        Methods & Params (B) & Inference FLOPs (T) & Inference Time (s) \\
        \midrule
        pixelSplat   & 0.119 & 0.764 & 0.152 \\
        MVSplat      & 0.012 & 0.170 & 0.059 \\
        NoPoSplat    & 0.612 & 0.405 & 0.042 \\
        SelfSplat    & 0.081 & 0.491 & 0.101 \\
        PF3plat      & 0.394 & 2.164 & 1.171 \\
        SPFSplat     & 0.616 & 0.405 & 0.044 \\
        SPFSplatV2   & 0.613 & 0.405 & 0.043 \\
        SPFSplatV2-L & 1.223 & 0.610 & 0.075 \\
        \bottomrule
    \end{tabular}
\end{table}

%% file: table/training_efficiency.tex
        


\begin{table}[t]
\caption{Comparison of Training Efficiency on an NVIDIA A100 GPU. }
\setlength{\tabcolsep}{4pt}
\label{tab:training_efficiency}
    \centering
    \begin{tabular}{lcccc}
        \toprule
        Methods  & Training FLOPs (T) & Training Time (s) & Mem. (GB) \\
        \midrule
        SelfSplat & 0.491 & 0.122 & 6.602 \\
        PF3plat & 2.164  & 0.633 & 15.043\\
        SPFSplat &  0.582 & 0.110 & 5.634 \\
        SPFSplatV2   & 0.515  & 0.082 & 4.891 \\
        SPFSplatV2-L  & 0.849  & 0.143  & 7.044 \\
        \bottomrule
    \end{tabular}

\end{table}

%% file: table/component_ablation.tex
\begin{table*}[ht]
\caption{\revision{Component ablations on RE10K. NVS$^\ast$ denotes novel view synthesis evaluated with pose alignment. }}
\label{tab:ablation}
\setlength{\tabcolsep}{5.5pt}
\centering
\begin{tabular}{l ccc ccc ccc ccc}
\toprule
\multirow{2}{*}{Config.} &
\multicolumn{3}{c}{\textbf{NVS}} &
\multicolumn{3}{c}{\textbf{NVS$^\ast$}} &
\multicolumn{3}{c}{\textbf{Pose}} &
\multicolumn{3}{c}{\textbf{Pose (PnP)}} \\
\cmidrule(lr){2-4}
\cmidrule(lr){5-7}
\cmidrule(lr){8-10}
\cmidrule(lr){11-13}
&
PSNR$\uparrow$ & SSIM$\uparrow$ & LPIPS$\downarrow$ &
PSNR$\uparrow$ & SSIM$\uparrow$ & LPIPS$\downarrow$ &
5$^\circ$ $\uparrow$ & 10$^\circ$ $\uparrow$ & 20$^\circ$ $\uparrow$ &
5$^\circ$ $\uparrow$ & 10$^\circ$ $\uparrow$ & 20$^\circ$ $\uparrow$ \\
\midrule

SPFSplatV2
& \textbf{25.693} & \textbf{0.853} & \textbf{0.149}
& \textbf{26.157} & \textbf{0.861} & \textbf{0.146}
& \textbf{0.638} & \textbf{0.776} & \textbf{0.863}
& \textbf{0.641} & \textbf{0.777} & \textbf{0.864} \\

w/o learnable pose token
& 25.636 & 0.851 & 0.150
& 26.098 & 0.859 & 0.147
& 0.634 & 0.772 & 0.859
& 0.632 & 0.770 & 0.858 \\

w/o masked attn. and pose token
& 25.484 & 0.847 & 0.153
& 25.845 & 0.852 & 0.152
& 0.617 & 0.755 & 0.845
& 0.613 & 0.754 & 0.845 \\

w/o intrinsics embedding
& 24.998 & 0.834 & 0.157
& 25.597 & 0.847 & 0.153
& 0.546 & 0.716 & 0.829
& 0.581 & 0.738 & 0.841 \\

w/o reprojection loss
& 19.818 & 0.651 & 0.280
& 22.013 & 0.751 & 0.244
& 0.023 & 0.144 & 0.393
& 0.015 & 0.064 & 0.197 \\

\bottomrule
\end{tabular}
\end{table*}

%% file: figure/ablation.tex
\begin{figure}[t]
    \centering
    \includegraphics[width=1.0\linewidth]{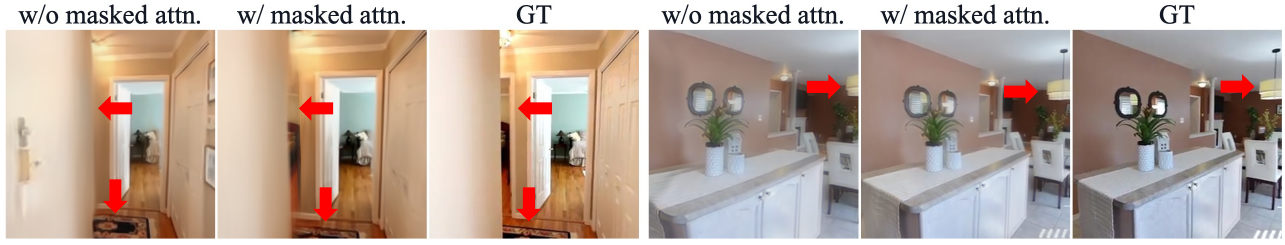}
   \caption{Ablation on masked attention. Masked attention improves the alignment between reconstructed 3D geometry and predicted target poses, reducing pose drift and producing more geometrically consistent reconstructions.}
    \label{fig:ablation}
\end{figure}

%% file: figure/ablation_multiview_dropout.tex
\begin{figure}[t]
    \centering
    \includegraphics[width=0.7\linewidth]{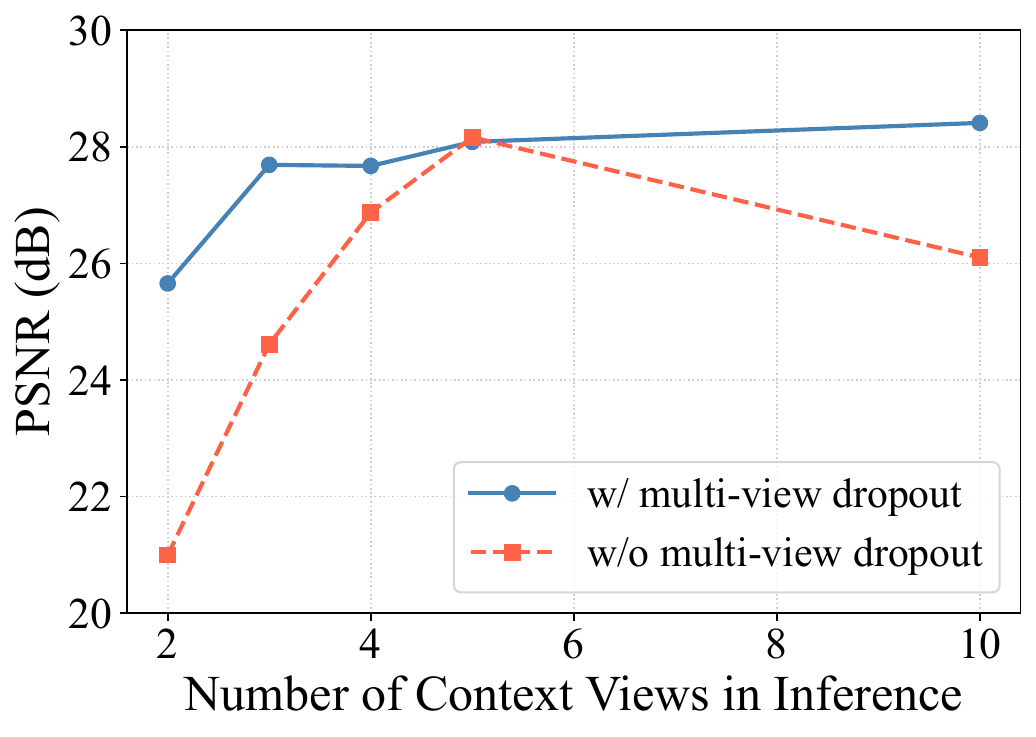}
    \caption{\revision{Ablation of multi-view dropout on RE10K with 5-view training inputs.}}
    \label{fig:multiview_dropout}
\end{figure}

%% file: table/gt_poses_ablation.tex
\begin{table}[t]
\footnotesize
\caption{Ground-truth poses ablations on RE10K.}
\label{tab:gt_poses_ablation}
\setlength{\tabcolsep}{3.2pt}
    \centering
    \begin{tabular}{l ccc ccc}
        \toprule
        \multirow{2}{*}{\textbf{Method}} & \multicolumn{3}{c}{\textbf{NVS$^\ast$}} & \multicolumn{3}{c}{\textbf{Pose}} \\
        \cmidrule(lr){2-4} \cmidrule(lr){5-7}  
        & PSNR$\uparrow$ & SSIM$\uparrow$ & LPIPS$\downarrow$ &  5$^\circ$ $\uparrow$ & 10$^\circ$ $\uparrow$ & 20$^\circ$ $\uparrow$  \\
        \midrule
        (a) SPFSplatV2 (Ours) & \textbf{26.157} & \textbf{0.861} & \textbf{0.146} & \underline{0.638} & \underline{0.776} & \underline{0.863}  \\
        (b) render with gt pose & 25.033 & 0.838 & 0.160 & 0.571 & 0.727 & 0.833 \\
        (c) w/ gt pose loss & \underline{25.910} & \underline{0.860} & \underline{0.150} & \textbf{0.693} & \textbf{0.814} & \textbf{0.889}\\
        \bottomrule
    \end{tabular}
\end{table}

%% file: figure/phone_image.tex
\begin{figure}[t]
    \centering
    \includegraphics[width=\linewidth]{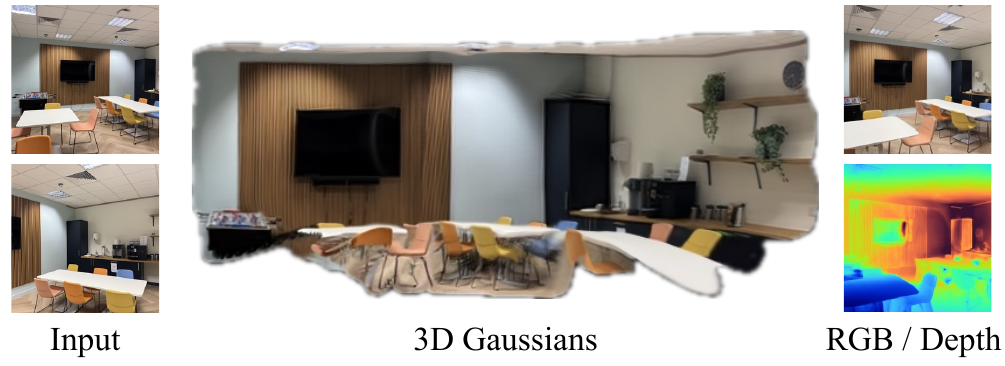}
    \caption{3D Gaussians, rendered images and depths from phone images. }
    \label{fig:phone_vis}
\end{figure}

%% file: sec/5_conclusion.tex
\section{Limitations and Future Work}
Our method can be trained without ground-truth poses and scales effectively to large datasets, opening the possibility for future work to exploit more diverse data to further improve pose estimation and generalization. Nonetheless, it still benefits from the priors provided by supervised models such as MASt3R and VGGT, as evidenced by the performance drop when training from random initialization. Furthermore, since our approach is not generative, it cannot reconstruct unseen regions with high-fidelity textures. Incorporating generative models is a promising direction to address this limitation.


\input{figure/failure_cases}

\section{Conclusion}
\label{sec:conclusion}
This paper presents SPFSplatV2, a self-supervised pose-free framework for 3D Gaussian splatting from sparse unposed views. By jointly optimizing camera poses and 3D Gaussian primitives through a unified backbone with masked attention, our approach achieves efficient and stable training as well as strong geometric consistency without requiring ground-truth poses. A reprojection loss is also incorporated with the conventional rendering loss to enforce pixel-aligned Gaussians. Extensive experiments on multiple datasets demonstrate that SPFSplatV2 and its larger variant SPFSplatV2-L establish new state-of-the-art results in novel view synthesis and cross-dataset generalization, \revision{while achieving strong relative pose estimation performance}, even under challenging conditions of extreme viewpoint change and limited overlap. Importantly, the framework’s independence from ground-truth poses underscores its scalability to large and diverse real-world datasets, paving the way for future advances in scalable and generalizable 3D reconstruction.

%% file: figure/failure_cases.tex
\begin{figure}[t]
    \centering
    \includegraphics[width=1.0\linewidth]{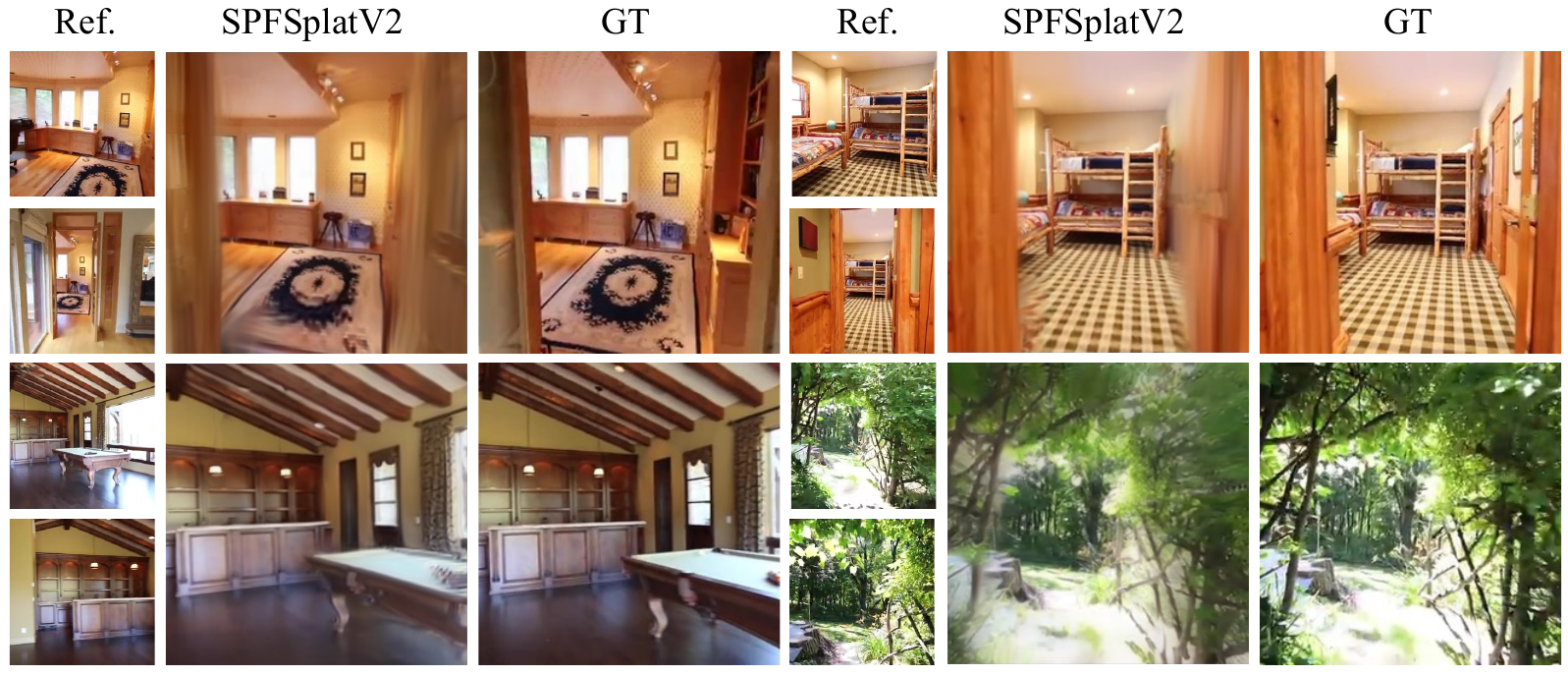}
    \caption{Failure cases of SPFSplatV2.  Blurriness and artifacts occur in occluded or texture-less regions and under extreme viewpoint changes.}
    \label{fig:bad_cases}
\end{figure}